\documentclass{article}
\usepackage[colorlinks,allcolors=blue]{hyperref}
\usepackage{graphicx}
\usepackage{geometry}
\usepackage{amsmath,amsthm,amssymb}
\usepackage{parskip}
\usepackage[noabbrev,nameinlink]{cleveref}
\usepackage[inline]{enumitem}
\usepackage{subcaption}
\usepackage{booktabs}
\usepackage{pgfplots}
\usepackage{todonotes}
\usepackage{array}
\usepackage{authblk}
\usepackage{algpseudocode,algorithm}
\usepackage{float}
\usepackage{mathtools}
\usepackage{natbib}

\numberwithin{equation}{section}
\numberwithin{algorithm}{section}

\newtheorem{theorem}{Theorem}[section] 
\newtheorem{corollary}[theorem]{Corollary}
\newtheorem{proposition}[theorem]{Proposition}
\newtheorem{lemma}[theorem]{Lemma}
\theoremstyle{definition}
\newtheorem{definition}[theorem]{Definition}

\newcommand{\norm}[1]{\left\|#1\right\|}

\newcommand{\inprod}[2]{\left\langle #1, #2\right\rangle}
\newcommand{\ie}{\textit{i.e.,\,}}

\newcommand{\ident}{\mathrm{Id}}
\newcommand{\real}{\mathbb R}
\newcommand{\codeurl}{\href{https://github.com/peck94/robust-width-defense}{this URL}}
\newcommand{\defect}{\mathcal E}

\DeclareMathOperator*{\expect}{\mathbb E}

\pgfplotsset{compat=1.18}
\usetikzlibrary{fit}

\newcounter{mysubequations}

\title{\textsc{Robust width:}\\A lightweight and certifiable adversarial defense}

\author[1,2]{Jonathan Peck}
\author[3]{Bart Goossens}

\affil[1]{Department of Applied Mathematics, Computer Science and Statistics, Ghent University}
\affil[2]{Data Mining and Modeling for Biomedicine, VIB Inflammation Research Center}
\affil[3]{Department of Telecommunications and Information Processing, imec/IPI/Ghent University}

\date{}

\begin{document} 

\maketitle

\begin{abstract}
\noindent Deep neural networks are vulnerable to so-called \textit{adversarial examples}: inputs which are intentionally constructed to cause the model to make incorrect predictions or classifications. Adversarial examples are often visually indistinguishable from natural data samples, making them hard to detect. As such, they pose significant threats to the reliability of deep learning systems. In this work, we study an adversarial defense based on the robust width property (RWP), which was recently introduced for compressed sensing. We show that a specific input purification scheme based on the RWP gives theoretical robustness guarantees for images that are approximately sparse. The defense is easy to implement and can be applied to any existing model without additional training or finetuning. We empirically validate the defense on ImageNet against $L^\infty$ perturbations at perturbation budgets ranging from $4/255$ to $32/255$. In the black-box setting, our method significantly outperforms the state-of-the-art, especially for large perturbations. In the white-box setting, depending on the choice of base classifier, we closely match the state of the art in robust ImageNet classification while avoiding the need for additional data, larger models or expensive adversarial training routines. Our code is available at \codeurl.
\end{abstract}


\section{Introduction}
\label{sec:intro}
\begin{figure}[ht]
    \begin{center}
        \begin{subfigure}[t]{\textwidth}
            \centering\includegraphics[width=.25\textwidth]{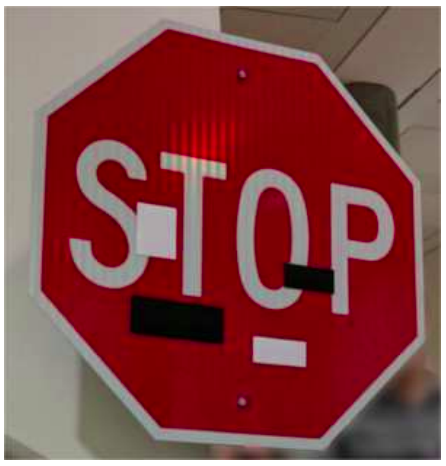}
            \caption{An adversarial stop sign~\citep{eykholt2018robust}. This sign will be incorrectly identified as a speed limit instead of a stop sign.}
            \label{fig:examples_stop}
        \end{subfigure}

        \begin{subfigure}[t]{\textwidth}
            \centering\includegraphics[width=\textwidth]{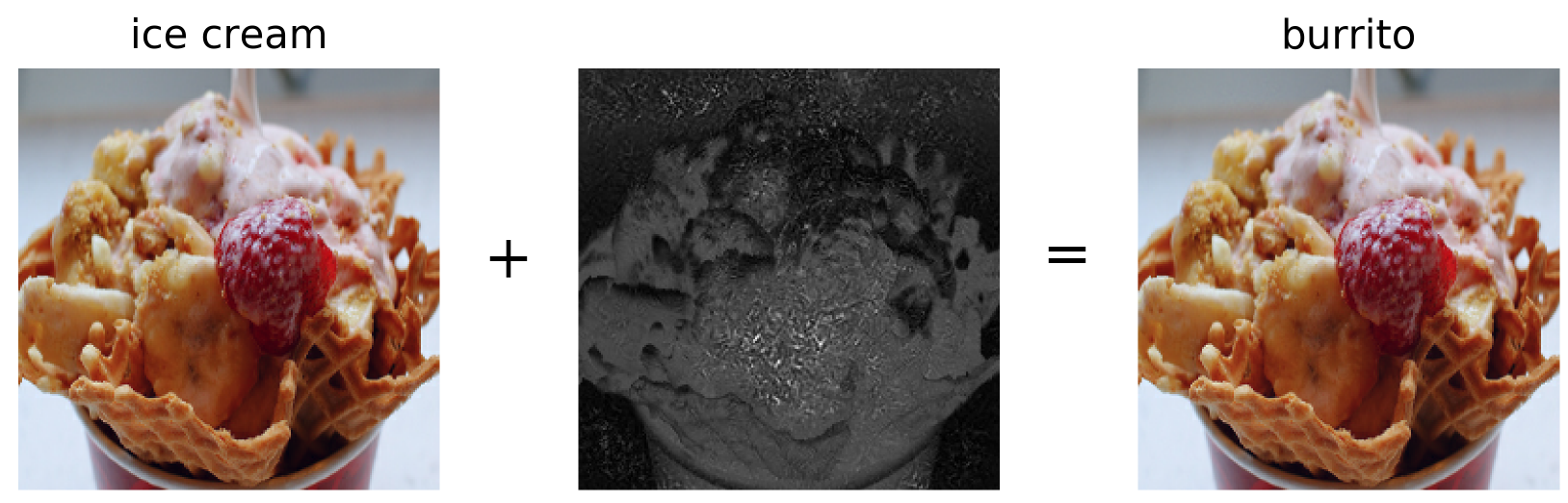}
            \caption{An imperceptible adversarial perturbation. The image on the right will be incorrectly labeled as ``burrito,'' whereas the perceptually indistinguishable original image on the left is correctly labeled as ``ice cream.''}
            \label{fig:examples_burrito}
        \end{subfigure}
    \end{center}
    \caption{Examples of adversarial perturbations in image recognition.}
    \label{fig:examples}
\end{figure}

Deep neural networks (DNNs) are known to be vulnerable to \textit{adversarial perturbations}. These are perturbations which have no significant effect on natural inputs, but which can cause state-of-the-art DNNs to output erroneous predictions~\citep{biggio2018wild}. \Cref{fig:examples} shows some examples of adversarial perturbations in the domain of image recognition. In \cref{fig:examples_stop}, black-and-white stickers are added to a stop sign in a way humans would consider harmless graffiti. However, classifiers trained for traffic sign recognition may mistake the stop sign for a speed limit sign. In \cref{fig:examples_burrito}, we add imperceptible perturbations to a sample from the ImageNet data set~\citep{deng2009imagenet} such that a standard ResNet50 model~\citep{he2016deep} incorrectly classifies it as a burrito despite the initially correct prediction of ice cream.

Generating adversarial perturbations that reliably fool highly accurate DNNs turns out to be surprisingly easy, even when little to no information about the target model is known~\citep{papernot2017practical}. On the other hand, making DNNs robust to adversarial perturbations while maintaining high accuracy is extremely difficult: despite over a decade of research on this problem, we are still far from satisfactory solutions~\citep{peck2024introduction}. At this time, the most popular defenses against adversarial attacks are adversarial training~\citep{qian2022survey,madry2017towards,goodfellow2014explaining} and randomized smoothing~\citep{cohen2019certified}. Adversarial training (AT) essentially uses adversarial perturbations as data augmentations, perturbing the training samples using an adversarial attack while the model is being trained. Though it can result in very robust models, AT methods tend to be slow since they require running an adversarial attack on each minibatch of training. They also tend to lack theoretical robustness guarantees, so any robustness these models have can only be verified empirically. Randomized smoothing (RS) requires no retraining of the model and enjoys theoretical guarantees. It takes a classifier $f$ and ``smooths'' it via the transformation
\begin{equation}\label{eq:rs}
    \hat{f}(x) = \arg\max_y~\Pr[f(x + \eta) = y].
\end{equation}
The randomness is taken over $\eta$, a zero-mean Gaussian random vector with tunable isotropic variance $\sigma^2$. In practice, $\hat{f}(x)$ is approximated by taking a large number of samples and aggregating using a majority vote. RS can certify local robustness around a given point $x$ according to the following formula:
\begin{equation}\label{eq:dsrob}
    \rho(x) = \frac{\sigma}{2}\left( \Phi^{-1}(p_1) - \Phi^{-1}(p_2) \right),
\end{equation}
where $\sigma$ is the standard deviation of the Gaussian noise, $\Phi^{-1}$ is the inverse CDF of the standard Gaussian distribution and $p_1, p_2$ are the two highest predicted probabilities. In case \eqref{eq:rs} is solved exactly, RS guarantees that no additive perturbation with $L^2$ norm less than $\rho(x)$ can change the output of the smoothed classifier $\hat{f}$. In practice, using sufficiently many samples for $\eta$, this bound can be certified with high probability. Although this result only holds for perturbations measured in $L^2$ norm, variations of RS exist which apply to other settings~\citep{yang2020randomized,levine2020wasserstein}.

Recent improvements on the basic randomized smoothing algorithm have focused on the inclusion of a denoiser, leading to a variant known as \textit{denoised smoothing}~\citep{salman2020denoised}:
\begin{equation}\label{eq:ds}
    \hat{f}(x) = \arg\max_y~\Pr[f(D(x + \eta)) = y].
\end{equation}
Here, $D: \mathcal X \to \mathcal X$ is a generative model trained to denoise the given images. Denoised smoothing (DS) has the same theoretical guarantees as RS, but it can maintain a higher clean accuracy thanks to the denoising step which reduces the influence of the Gaussian noise. This also allows for higher values of $\sigma$, which improves the theoretical bound $\rho(x)$ provided the denoiser $D$ is sufficiently powerful.

DS has emerged as the preferred method for obtaining certified adversarial robustness. Compared to almost all alternative defense methods, DS is exceedingly easy to understand and implement, and its robustness certificates are straightforward to compute. It therefore holds the promise of being an efficient ``plug-and-play'' adversarial defense~\citep{carlini2023certified}. At this time, however, the reality is slightly more nuanced. First, DS can be computationally expensive due to the large number of samples that need to be taken to certify robustness with sufficient certainty. The denoiser $D$ is typically a DNN (often a diffusion model), so this limits the usefulness of DS in low-resource environments.\footnote{For example, the robustness gains reported by \citet{carlini2023certified} are obtained by prepending a 552 M parameter diffusion model to a 305 M parameter base classifier. Assuming 64-bit precision, it would take 7 GB to simply load such a model into memory.} Moreover, DS crucially relies on a strong denoiser $D$ to obtain good robustness certificates. In modern implementations, these denoisers take the form of diffusion models trained on millions of data samples~\citep{carlini2023certified}. This is feasible when off-the-shelf models exist for the data set in question, or when sufficient data is available to perform fine-tuning or train novel denoisers from scratch. In many important applications, however, data is not abundant and off-the-shelf denoisers may not be available~\citep{alzubaidi2023survey}. This is often the case in medical imaging settings, where reliability of the models is paramount and robustness is an important desideratum~\citep{finlayson2019adversarial,hirano2021universal,ma2021understanding}. It is therefore necessary to develop adversarial defenses which improve robustness without sacrificing performance with respect to other metrics (such as standard accuracy, false positive rate, etc) while avoiding the need for massive data sets, extremely large models and time-consuming inference procedures.

In this work, we aim to address the above problems by designing an adversarial defense similar to DS but which replaces the denoiser $D$ by a lightweight algorithm that needs no training. This allows our defense to be deployed in settings where computational resources are limited or data is too scarce to train effective denoisers.

Our approach makes use of the robust width property (RWP) introduced by \citet{cahill2021robust}, which is a geometric criterion that determines whether the solution to a convex optimization problem offers a precise solution to an underdetermined noise-affected linear system. It guarantees that the error bound is small if the image is approximately sparse and the perturbation is small. We show that the RWP can be used to construct a defense that is robust to adversarial perturbations, by relating the RWP error bound to the robustness of the classifier. This will allow us to certify the robustness of the RWP-based adversarial defense. Although in this paper, we consider adversarial robustness in the $L^2$ and $L^\infty$ sense, the RWP-based defense can be extended to more general $L^p$-norms (see e.g., \cite{yu2019stable}).

We evaluate our defense method in both the black-box setting (where the adversary does not have direct access to the model) as well as the white-box setting (where the adversary has full access to the model). In the black-box setting, we evaluate against transfer attacks using surrogate models as well as direct attacks against the defended models themselves. The results show that our method significantly improves on state-of-the-art robustness in the black-box setting, especially for large perturbation budgets. In the white-box setting, depending on the choice of base classifier, we closely match state-of-the-art robust accuracy without the need for adversarial retraining of the models.

\subsection*{Our contributions}
To summarize, we make the following contributions: 
\begin{enumerate}
    \item We derive novel robustness guarantees based on compressed sensing using the robust width property. Our robustness certificates depend on the assumption that the data manifold consists of vectors which are \emph{approximately} sparse in some appropriate basis. Under such a sparsity assumption, we can give bounds for any classifier that depend \textit{linearly} on the size of the perturbation as well as the distance of the natural samples to the data manifold.

    \item We empirically validate the robustness of the proposed method on the ImageNet data set~\citep{deng2009imagenet} against APGD and Square attacks~\citep{andriushchenko2020square,croce2020reliable}. In the black-box setting, we significantly outperform existing defenses, especially at large perturbation budgets. In the white-box setting, we closely match the state of the art using pre-trained standard (non-robust) models, without needing to resort to adversarial training or larger data sets.
\end{enumerate}
Our code is publicly available at \codeurl.

\subsection*{Outline}
This paper is organized as follows. In \cref{sec:prelim}, we introduce the necessary background on robustness of classifiers, compressed sensing and the RWP. \Cref{sec:related} surveys some related work in robustness based on sparsity assumptions. In \cref{sec:RWP_defense}, we show that the RWP can be used to construct a defense that is provably robust to adversarial perturbations. In \cref{sec:experiments}, we empirically evaluate the defense on the ImageNet data set using a variety of models and compare the performance to the state-of-the-art at the time of this writing as reported by the RobustBench leaderboard.\footnote{\url{https://robustbench.github.io/}. Accessed 2024-05-25.} Finally, \cref{sec:conclusion} concludes this paper and provides some avenues for future research.

\section{Preliminaries}
\label{sec:prelim}
We consider the supervised learning setting where the goal is to find a classifier $f: \mathcal V \to \mathcal W$ which maps input samples from an inner product space $\mathcal V$ to labels in a discrete set $\mathcal W = \{ 1, \dots, K \}$. In most applications, the input space $\mathcal V$ is a compact subset of $\real^n$. As an inner product space, $\mathcal V$ is equipped both with an inner product $\inprod{\cdot}{\cdot}$ and a norm $\norm{x}_2 = \sqrt{\inprod{x}{x}}$.

\subsection{Adversarial robustness}
It is well-known that state-of-the-art machine learning classifiers are highly vulnerable to small perturbations of their inputs, even when their accuracy on natural data is very high~\citep{peck2024introduction,biggio2018wild,szegedy2013intriguing}. Specifically, when the input data have the structure of a vector space, then for any input $x \in \mathcal V$ it is generally easy to find some $\delta \in \mathcal V$ such that $\norm{\delta}$ is very small but $f(x) \neq f(x + \delta)$. These perturbations $\delta$ are called \textit{adversarial perturbations}. They are typically constructed by solving the following optimization problem:
\begin{equation}\label{eq:adv_opt}
    \delta^\star(x) = \arg\min_{\delta \in \mathcal V}~\norm{\delta} \mbox{ subject to } f(x) \neq f(x + \delta).
\end{equation}
Here, $\norm{\cdot}$ can be any norm on $\mathcal V$. The most common choices in the literature are $L^2$ and $L^\infty$ norms.

Various practical algorithms or \textit{adversarial attacks} have been proposed to efficiently find solutions to \eqref{eq:adv_opt}, such as fast gradient sign~\citep{goodfellow2014explaining} and projected gradient descent~\citep{madry2017towards}. We thus define our notion of robustness as invariance to sufficiently small additive perturbations of the input:

\begin{definition}[Robustness of a classifier on a set]
\label{def:sample_robustness}
A classifier $f: \mathcal V \to \mathcal W$ is called $\tau$-robust on a sample set $S \subseteq \mathcal V$ if
$$
    f(v) = f(v + \delta)
$$
for all $v \in S$ and $\delta \in \mathcal V$ with $\norm{\delta}_2 < \tau$. The robustness bound $\tau$ is \textit{tight} if there is no $\tau' > \tau$ such that $f$ is $\tau'$-robust on $S$.
\end{definition}

The robustness of a classifier is a measure of how much the classifier's output changes when the input is perturbed. A classifier is $\tau$-robust on a sample set $S$ if the classifier's output does not change when the input is perturbed by a small $\delta$. In case the sample set is a singleton, the robustness bound $\tau$ indicates local robustness in the particular point.

\subsection{Compressed sensing}\label{sec:cs}
Compressed sensing (CS) is a field of signal processing which attempts to reconstruct signals based on compressed\footnote{This is where the name \textit{compressed} sensing comes from: it derives from the fact that we are acquiring signals which have been compressed beforehand. In CS, we typically never have access to raw uncompressed signals.} measurements which can be noisy. Formally, CS assumes the original signal $x^\natural$ comes from a Hilbert space $\mathcal H$. The acquisition process is modeled by a linear operator $\Phi: \mathcal H \to \mathcal H'$, where $\mathcal H'$ is another Hilbert space. The dimension of $\mathcal H'$ will typically be much smaller than the dimension of $\mathcal H$, so that $\Phi$ provides us with a compressed representation of $x^\natural$. Our observations now take the form
$$
    y = \Phi x^\natural + e
$$
where $e \in \mathcal H'$ is a noise vector satisfying $\norm{e}_2 \leq \varepsilon$ with known $\varepsilon$. The goal of CS is to reconstruct $x^\natural$ based on this noisy compressed observation $y$. To this end, we require the notion of a \textit{frame}:

\begin{definition}[Frame]
Given an inner product space $\mathcal V$, a set of vectors $\{u_n\}_{n \in \mathbb N}$ in $\mathcal V$ is called a \textit{frame} if it satisfies the frame condition:
$$
    A\norm{v}^2 \leq \sum_{n \in \mathbb N}|\inprod{v}{u_n}|^2 \leq B\norm{v}^2
$$
for constants $A, B > 0$ and every $v \in \mathcal V$. The frame is \textit{tight} if $A = B$.
\end{definition}

Frames can be equipped with an \textit{analysis} operator and a \textit{synthesis} operator. The analysis operator of a frame maps vectors $v \in \mathcal V$ to vectors of their corresponding frame coefficients $\inprod{v}{u_1}, \inprod{v}{u_2}, \dots$ The synthesis operator performs the inverse operation, mapping vectors of frame coefficients to vectors in $\mathcal V$.

The basic problem in CS is the reconstruction of the original signal $x^\natural$ based on the observation $y$. This is typically formulated as the following optimization problem:
\begin{equation}\label{eq:cs_opt}
    \hat{x} = \arg\min_{x \in \mathcal{H}}~\norm{\Psi x}_1 \mbox{ subject to } \norm{\Phi x - y}_2 \leq \varepsilon.
\end{equation}
Here, $\Psi$ is the analysis operator of a frame chosen such that natural signals $x^\natural$ will have small norm $\norm{\Psi x^\natural}_1$. CS therefore works best when the natural signals $x^\natural$ have sparse representations in some known frame. Common examples include the Fourier and wavelet transforms, which are well-suited for problems such as the reconstruction of magnetic resonance images~\citep{ye2019compressed,ouahabi2013review}. Many algorithms exist for solving the optimization problem \eqref{eq:cs_opt}. We refer the reader to existing surveys such as \citet{foucart2015mathematical} for a detailed overview.

\subsection{The robust width property}
\citet{cahill2021robust} originally introduced the \textit{robust width property} (RWP) as a geometric criterion to delineate instances where the solution to a convex optimization problem offers a precise approximate solution to an underdetermined noise-affected linear system. The notion of \textit{robust width} applies to a wide variety of Hilbert spaces, which are referred to as \textit{compressed sensing spaces} or \textit{CS spaces}:

\begin{definition}[CS space]
A CS space with bound $L$ is a tuple $(\mathcal{H}, \mathcal{A}, \norm{\cdot}_\sharp)$ where $\mathcal{H}$ is a Hilbert space, $\mathcal{A} \subseteq \mathcal{H}$ is the solution space and $\norm{\cdot}_\sharp$ is a norm on $\mathcal{H}$. Furthermore, the following properties must hold:
\begin{itemize}
    \item $0 \in \mathcal{A}$
    \item For every $a \in \mathcal{A}$ and $z \in \mathcal{H}$, there exists a decomposition $z = z_1 + z_2$ such that
    $$
        \norm{a + z}_\sharp = \norm{a}_\sharp + \norm{z_1}_\sharp
    $$
    with $\norm{z_2}_\sharp \leq L\norm{z}_2$.
\end{itemize}
\end{definition}

A typical example of a solution space $\mathcal{A}$ is the subset of vectors in $\mathcal{H}$ with a bounded $\sharp$-norm:
\begin{equation}
\label{eq:solution_space}
    \mathcal{A} = \{ x \in \mathcal{H} \mid \norm{x}_\sharp \leq T \}
\end{equation}
for some bound $T > 0$. 

Note that a CS space generalizes the theory of compressed sensing we reviewed in \cref{sec:cs}. Specifically, the sparsity norm $\norm{\cdot}_\sharp$ is typically defined as
$$
    \norm{x}_\sharp = \norm{\Psi x}_1
$$
where $\Psi$ is an analysis operator of a suitable frame. Using \eqref{eq:solution_space}, the solution space $\mathcal A$ then consists of all vectors $x \in \mathcal H$ which have small norm in the frame defined by $\Psi$.

Given a CS space $(\mathcal{H}, \mathcal{A}, \norm{\cdot}_\sharp)$ and a linear operator $\Phi: \mathcal{H} \to \mathcal{H}'$, we can define the reconstruction of $x^\natural$ from the observation $y$ as
\begin{equation}\label{eq:delta}
    \Delta_\varepsilon(y) = \arg\min_{x \in \mathcal{H}}~\norm{x}_\sharp \mbox{ subject to } \norm{\Phi x - y}_2 \leq \varepsilon.
\end{equation}
Optimization problems of this form are well-studied in the literature, and many efficient algorithms have been proposed to solve them; see, for example, \citet{daubechies2004iterative} and \citet{foucart2015mathematical}.

The question then arises to what extent solutions to \eqref{eq:delta} can be accurate. It is no surprise that accurate reconstructions cannot be guaranteed without assumptions on $\Phi$ and $\Psi$. The most widely used assumption to this end is the so-called \textit{restricted isometry property} (RIP):

\begin{definition}[Restricted isometry property]
Let $J \in \mathbb N$ and $\delta > 0$. A linear operator $\Phi: \mathcal{H} \to \mathcal{H}'$ satisfies the $(J,\delta)$-restricted isometry property (RIP) if
$$
    (1 - \delta)\norm{u}_2^2 \leq \norm{\Phi u}_2^2 \leq (1 + \delta)\norm{u}_2^2
$$
for every $J$-sparse $u \in \mathcal{H}$.
\end{definition}

The RIP guarantees small reconstruction error in \eqref{eq:delta} provided the signals $x^\natural$ are \textit{exactly} sparse. Variations of the RIP have been proposed in order to bound the reconstruction error in cases where the signal is only \textit{approximately} sparse. The robust width property (RWP) is one such assumption: it characterizes to what extent the null space of a linear operator intersects with the unit $\sharp$-ball $B_\sharp = \{ x \in \mathcal{H} \mid \norm{x}_\sharp \leq 1 \}$.

\begin{definition}[Robust width property]
Let $\alpha, \rho > 0$. A linear operator $\Phi: \mathcal{H} \to \mathcal{H}'$ satisfies the $(\rho, \alpha)$-robust width property (RWP) if
$$
    \norm{\Phi u}_2 \leq \alpha\norm{u}_2 \implies \norm{u}_2 < \rho\norm{u}_\sharp
$$
for every $u \in \mathcal{H}$.
\end{definition}

With respect to a chosen frame $\Psi$, the RWP condition can be formulated as
$$
    \norm{\Phi u}_2 \leq \alpha\norm{u}_2 \implies \norm{u}_2 < \rho\norm{\Psi u}_1.
$$
Hence the RWP relates the sensing operator $\Phi$ to the chosen frame via the analysis operator $\Psi$.

The following theorem from \cite{cahill2021robust} gives an error bound on the reconstruction for solutions $\Delta_\varepsilon(\Phi x^\natural + e)$ to the above convex optimization problem \eqref{eq:delta}:
\begin{theorem}[\cite{cahill2021robust}]\label{thm:rwp}
Let $(\mathcal{H}, \mathcal{A}, \norm{\cdot}_\sharp)$ be a CS space with bound $L$ and let $\Phi: \mathcal{H} \to \mathcal{H}'$ be a linear operator satisfying the $(\rho, \alpha)$-RWP. Let $x^\natural \in \mathcal{H}$, $e \in \mathcal{H}'$ with $\norm{e}_2 \leq \varepsilon$ for $\varepsilon > 0$. It then holds that
$$
    \norm{\Delta_\varepsilon(\Phi x^\natural + e) - x^\natural}_2 \leq \frac{2\varepsilon}{\alpha} + 4\rho\inf_{a \in \mathcal{A}}\norm{x^\natural - a}_\sharp
$$
provided that $\rho \leq \frac{1}{4L}$.
\end{theorem}

Here, the error bound consists of two terms: a first term is proportional to the perturbation $\varepsilon$ and a second term is proportional to $\inf_{a \in \mathcal{A}}\norm{x^\natural - a}_\sharp$, which we refer to as the \textit{sparsity defect} of the signal or image $x^\natural$. This notion will appear often enough in this work to justify a separate definition:

\begin{definition}[Sparsity defect]
Let $\mathcal S = (\mathcal H, \mathcal A, \norm{\cdot}_\sharp)$ be a CS space. The \textit{sparsity defect} of $x \in \mathcal H$ is defined as
$$
    \defect_{\mathcal S}(x) = \inf_{a \in \mathcal A}~\norm{x - a}_\sharp.
$$
We will omit the subscript $\mathcal S$ whenever it is clear from the context.
\end{definition}

The sparsity defect $\defect(x^\natural)$ quantifies the extent to which  $x^\natural$ is not sparse in the space $\mathcal{A}$. Such defects can occur due to misspecification of the space $\mathcal A$ or it can be caused by noise (random or adversarial). The RWP guarantees that the error bound is small if the image is approximately sparse and the perturbation is small. This is an advantage over classical compressed sensing, which only gives recovery guarantees for exactly sparse signals or images.

We note that \cref{thm:rwp} is specific to perturbations $e$ which are bounded in $L^2$ norm. However, \citet{yu2019stable} have shown that the reconstruction guarantees proven by \citet{cahill2021robust} can be extended to arbitrary $L^p$ norms. Hence similar bounds can be expected to hold for $L^\infty$ perturbations, for instance.

\section{Related work}\label{sec:related}
The field of adversarial robustness has developed rapidly over the past decade since the seminal work of \citet{szegedy2013intriguing}. For a general overview of the field, we refer the reader to existing surveys such as \citet{biggio2018wild} and \citet{peck2024introduction}. Here, we will focus exclusively on adversarial defenses based on sparsity assumptions, as these are closest in spirit to this work.

The idea that sparsity plays an important role in adversarial robustness is not new. However, a distinction must be made between approaches which incorporate sparsity into the \textit{model} versus approaches which sparsify the \textit{data}. Sparsifying or \textit{pruning} the weights of a model is an established technique for improving not only computational and memory efficiency, but adversarial robustness as well~\citep{guo2018sparse,molchanov2017variational,hoefler2021sparsity,liang2021pruning}. These techniques require modification of the model during or after training. In this work, we focus on a ``plug-and-play'' defense which requires no modification of the underlying model. Of course, pruning may still be applied in conjunction with our method, so our work is in some sense orthogonal to pruning approaches.

Specifically, we develop a particular denoising scheme based on compressed sensing which exploits sparsity of the data itself. Similar defensive approaches have been explored in prior work. \citet{bhagoji2018enhancing} propose linear dimensionality reduction as an effective preprocessing step to defend classifiers against adversarial perturbation. Other sparsifying data transformations such as JPEG compression and bit-depth reduction have been explored by \citet{guo2017countering,das2017keeping}. However, the robustness gained from such transformations is often illusory due to a phenomenon known as \textit{gradient masking}~\citep{athalye2018obfuscated}. Gradient masking typically occurs when a model utilizes non-differentiable components when processing the data. If the gradient of the loss with respect to the input cannot be computed accurately, then gradient-based adversarial attacks can fail due to numerical errors rather than an inherent improvement of model robustness. Often, such models can still be attacked very effectively if one is aware of the masking phenomenon, either by approximating the non-differentiable components with differentiable ones or carrying out transfer attacks~\citep{croce2022evaluating}.

\begin{figure}
    \centering
    \begin{tikzpicture}[
    	rect/.style = {rectangle, solid, draw=black!100, fill=cyan!18, thin, minimum height=7.5mm, minimum width=11mm},
    	rect_smaller/.style = {rectangle, solid, draw=black!100, fill=cyan!18, thin, minimum height=7.5mm, minimum width=8.5mm},
    	circ/.style = {circle, solid, draw=black!100, fill=cyan!18, thin, minimum size=7mm},
    	outer/.style = {rounded corners=0.2cm, draw=black!100, dashed, inner sep = 2.2mm}
    	]
    	\node[] (x) {$x$};
    	\node[circ] (plus) [right = 5mm of x] {$+$};
    	\node[] (e) [above = 5mm of plus] {$e$};
    	\node[rect_smaller] (psit) [right = 5mm of plus] {$\Psi^T$};
    	\node[rect_smaller] (hk) [right = 3mm of psit] {$\mathcal H_K(\cdot)$};
    	\node[rect_smaller] (psi) [right = 3mm of hk] {$\Psi$};
    	\node[rect] (w) [right = 9mm of psi] {$f(\cdot)$};
    	\node[] (out) [right = 3mm of w] {$\hat{y}$};
    	\node[outer,fit = (psit) (hk) (psi), label=below:\footnotesize{Sparsifying front end}] (frontend) {};
    	\draw[->] (x) -- (plus);
    	\draw[->] (e) -- (plus); 
    	\draw[->] (plus) -- (psit); 
    	\draw[->] (psit) -- (hk);
    	\draw[->] (hk) -- (psi);
    	\draw[->] (psi) -- node[anchor=south]{$\hat{x}$} (w);
    	\draw[->] (w) -- (out);
	\end{tikzpicture}
    \caption{Diagram of the sparsifying front-end proposed by \citet{marzi2018sparsity}.}
    \label{fig:marzi}
\end{figure}
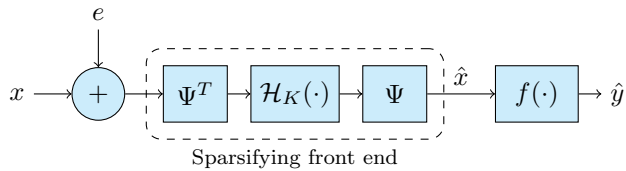

\citet{marzi2018sparsity} provide theoretical results showing that sparsifying data transformations can improve robustness against $L^\infty$ adversarial attacks. A diagram of their approach is shown in \cref{fig:marzi}. Here, $\Psi$ is an orthonormal basis such that $\norm{\Psi x}_0 \leq K$ for some $K > 0$, \ie the data vectors are assumed to be $K$-sparse in the basis $\Psi$. The function $\mathcal H_K$ is a hard thresholding operator which only retains the $K$ largest components of the input vector. The present work can be seen as an extension of the method proposed by \citet{marzi2018sparsity}, where recent results in the field of compressed sensing are integrated into the defense. In particular, our approach only requires \textit{approximate} sparsity in the chosen basis. Our theoretical results are also generally applicable to any classifier rather than only locally linear ones.

\citet{bafna2018thwarting,dhaliwal2019recovery} introduced the \textit{Compressive Recovery Defense} (CRD) framework, which is also based on compressed sensing and yields certifiable robustness bounds using the restricted isometry property (RIP). By comparison, our results are based on the more recent robust width property (RWP). The RWP is strictly weaker than the RIP, but the resulting bounds are more interpretable and depend on factors that are easier to control in practice. \citet{sulam2020adversarial} have similarly made use of the RIP in sparse coding to provide certifiable robustness as well as generalization bounds for the robust risk. The true robustness of these types of input purification methods has been contested in the past, with over-estimation of robustness being a common issue~\citep{tramer2020adaptive}. Therefore, our robustness evaluation is carried out according to the recommendations proposed by \citet{croce2022evaluating}. In particular, we perform extensive experiments to rule out obfuscated gradients. We also explicitly discuss circumstances where our results do not hold or only hold vacuously.

Other adversarial defenses in this vein make use of sparse transformations such as the discrete wavelet transform (DWT), discrete cosine transform (DCT) or discrete Fourier transform (DFT), based on the idea that adversarial perturbations mainly concentrate in high-frequency components of the input~\citep{sarvar2022defense,sinha2023adversarial,zhang2019adversarial}. It is important to note, however, that this hypothesis has been undermined by works such as \citet{maiya2021frequency} as well as the effectiveness of low-frequency adversarial attacks~\citep{guo2019simple}. Adversarial examples do not seem to reliably concentrate in either high or low-frequency components. Indeed, their structure is heavily dataset-dependent, and so we cannot expect to be robust by simply removing components within some fixed frequency range.

\section{RWP-based adversarial defense}
\label{sec:RWP_defense}
RWP-based compressed sensing allows to recover approximately sparse images from measurements corrupted by additive noise. Crucially, there are no assumptions on the noise vectors aside from boundedness in an appropriate norm, which is usually satisfied for adversarial perturbations. Hence it is natural to use the RWP in adversarial defense. In this section, we will show how to use the RWP to construct a defense that is robust to adversarial perturbations, by relating the RWP error bound to the robustness of the classifier. This will allow us to certify the robustness of the RWP-based adversarial defense, in the sense of \cref{def:sample_robustness}. Because sensing matrices in CS are typically stochastic, we will also consider the probabilistic robustness of this defense.

\subsection{Adversarial denoiser}
\label{sec:adv_denoiser}

First, we define an adversarial denoiser that is specifically designed to remove adversarial perturbations $\delta$ with a given budget $\varepsilon$, i.e., $\norm{\delta}_2 < \varepsilon$. 

\begin{definition}[Adversarial denoiser]
\label{def:adv_denoiser}
    Let $f : \mathcal V \to \mathcal W$ be a classifier. Then $g_\varepsilon : \mathcal V \to \mathcal V$ is an adversarial denoiser for $f$ on a sample set $S \subset \mathcal V$, with perturbation budget $\varepsilon$ and performance bound $C(\varepsilon) \geq 0$, if for all $v \in S$ and $\delta \in \mathcal V$,
    \begin{subequations}
        \begin{eqnarray}
            \label{eq:defense}
            (f \circ g_\varepsilon)(v) = f(v) \quad \mbox{and} \quad  \\
            \label{eq:defense2}
            \norm{\delta}_2 < \varepsilon
            \Rightarrow
            \norm{g_\varepsilon(v + \delta) - v}_2 \leq C(\varepsilon).
        \end{eqnarray}
    \end{subequations}
\end{definition}

The defense method is required to preserve the classification result on the input sample set. In addition, the method must satisfy a denoising performance constraint. Denoting $\ident_{\mathcal V}$ as the identity operator on $\mathcal V$, a trivial denoiser is $g_\varepsilon = \ident_{\mathcal V}$ with performance bound $C(\varepsilon) = \varepsilon$. However, we are interested in denoisers with $C(\varepsilon) < \varepsilon$ which improve the robustness of the classifier. 

\begin{definition}[Deterministic adversarial defense method]
    A deterministic adversarial defense method consists of a classifier $f: \mathcal V \to \mathcal W$ and an adversarial denoiser $g_\varepsilon: \mathcal V \to \mathcal V$ for $f$ with perturbation budget $\varepsilon$. The smoothed classifier is  given by
    $$
        f \circ g_\varepsilon: \mathcal V \to \mathcal W.
    $$
\end{definition}

\begin{figure}[ht]
    \centering
    \begin{tikzpicture}
      \draw[domain=-2:2,samples=50,color=black!50,thick] plot (\x, \x^5/10 - \x);
      \fill (1.3,-0.7) circle (0.6mm);
      \draw[dashed] (1.3,-0.7) circle[radius=1cm];
      \node at (1.3,-0.5) {$x$};
    \end{tikzpicture}
    \caption{A hypothetical scenario where RS can incorrectly change the predicted label of a given sample $x$. The solid black line is the decision boundary and the dashed circle around $x$ represents an area of high probability where RS will sample from, such as a 99\% interval. If the original classification of $x$ by the base classifier was correct, RS will increase robustness around $x$ but at the cost of lowering accuracy.}
    \label{fig:bdy}
\end{figure}
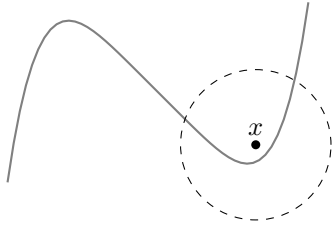

In general, it is impossible to construct a defense method without requiring some non-zero robustness of the original classifier. It is easy to see why: if a given sample $x$ lies sufficiently close to the decision boundary of a highly accurate classifier, then robustness around $x$ cannot be meaningfully improved without reducing accuracy. \Cref{fig:bdy} illustrates this problem in the context of randomized smoothing. This weakness of RS is widely known and is a subject of ongoing research~\citep{alfarra2022data,anderson2022certified,rumezhak2023rancer}. RS also tacitly assumes some baseline robustness of the underlying classifier in the form of $\Phi^{-1}(p_1) - \Phi^{-1}(p_2)$. This quantity directly depends on the probability margin $p_1 - p_2$ and hence can be translated into a robustness requirement based on the continuity properties of the classifier.

Therefore, we will start from a $\tau$-robust classifier $f$ and we will improve its robustness by applying an adversarial denoiser as a pre-processing step. Ideally, we wish that the smoothed classifier $f \circ g_\varepsilon$ be $\tau'$-robust for some $\tau' > \tau$.

\begin{definition}[Robustness gain of a deterministic adversarial defense method]
\label{def:robustness_gain}
    Let $f: \mathcal V \to \mathcal W$ be a classifier and let $g_\varepsilon : \mathcal V \to \mathcal V$ be an adversarial denoiser for $f$ in $S \subset \mathcal V$. If $f$ is $\tau$-robust in $S$ and $f \circ g_\varepsilon$ is $\tau'$-robust in $S$, where $\tau$ and $\tau'$ are both tight bounds (according to \cref{def:sample_robustness}), then the robustness gain in $S$ is $\tau'/\tau$.
\end{definition}

With this definition, the trivial adversarial denoiser $g_\varepsilon = \ident_{\mathcal V}$ has a robustness gain of 1. The following proposition establishes conditions under which the defense method can improve the robustness of a given classifier:

\begin{proposition}[An adversarial denoising method improving robustness]
\label{prop:adv_denoiser_robustness}
    Let $f: \mathcal V \to \mathcal W$ be a $\tau$-robust classifier and let $g_\varepsilon : \mathcal V \to \mathcal V$ be an adversarial denoiser for $f$ on $S \subset \mathcal V$ with performance bound $C(\varepsilon) = \varepsilon \kappa(\varepsilon) \leq \tau$, where $\kappa(\varepsilon)$ is an increasing function of $\varepsilon$. Then $f \circ g_\varepsilon$ is an $\varepsilon$-robust classifier for $S$. If additionally $\tau = \varepsilon \kappa(\varepsilon)$ is a tight robustness bound and $\kappa(\varepsilon) < 1$, then the robustness gain of the defense method in $S$ is at least $1/\kappa(\varepsilon)$.
\end{proposition}

\begin{proof}
    For a perturbation $\delta \in \mathcal V$ with $\norm{\delta}_2 < \varepsilon$, define $\delta' = g_\varepsilon(v + \delta) - v$. According to \cref{def:adv_denoiser}, $\norm{\delta'}_2 < C(\varepsilon) = \varepsilon \kappa(\varepsilon) \leq \tau$. Then for all $v \in S$,
    \begin{alignat*}{3}
        & v + \delta'  && = g_\varepsilon(v + \delta)  \\
        \Rightarrow & f(v + \delta')  && = (f \circ g_\varepsilon)(v + \delta) \\
        \Rightarrow & f(v) && = (f \circ g_\varepsilon)(v + \delta) \\
        \Rightarrow & (f \circ g_\varepsilon)(v)  && = (f \circ g_\varepsilon)(v + \delta)
    \end{alignat*}
    where we used the $\tau$-robustness of $f$ and \cref{def:adv_denoiser}. Consequently, $f \circ g_\varepsilon$ is $\varepsilon$-robust in $S$. If $C(\varepsilon) = \varepsilon \kappa(\varepsilon) = \tau$ and considering that $\varepsilon \kappa(\varepsilon)$ is not necessarily a tight bound for $\norm{\delta'}_2$, it follows that the robustness gain is at least $\varepsilon / \tau = 1/\kappa(\varepsilon)$.
\end{proof}

The above proposition relates the robustness bounds to the performance of the adversarial denoiser. It is necessary to make a distinction between tight and non-tight bounds $\tau$, because for a non-tight bound, the robustness gain cannot be measured, according to \cref{def:robustness_gain}. More specifically, if $\tau$ is under-estimated, a robustness improvement is possible even if the gain compared to this lower bound is less than one.

To proceed, we introduce the concept of a bounded denoiser, which is commonly used in the literature on plug-and-play denoising methods (see, e.g., \cite{chan2016plug}) and which will allow us to verify if a denoiser is an adversarial denoiser for a given $\tau$-robust classifier.
\begin{definition}[Bounded denoiser]
    A denoiser $g_\varepsilon: \mathcal V \to \mathcal V$ is $C$-bounded if $\norm{g_\varepsilon(v) - v}_2 \leq C$ for all $v \in \mathcal V$ and some $C > 0$.
\end{definition}

In practice most denoising methods are bounded if constrained to a pixel value interval such as $[0,1]$ or $[0,255]$. However, the bounds that we consider here are much smaller, typically on the order of the perturbation budget $\varepsilon$. Next, we show that a bounded denoiser can be used to construct an adversarial denoiser for a given classifier. 

\begin{corollary}
\label{cor:bounded_denoiser}
    Let $f: \mathcal V \to \mathcal W$ be a $\tau$-robust classifier on $S \subset \mathcal V$. If $g_\varepsilon$ is a $\tau$-bounded denoiser for a given $\varepsilon \geq 0$ and $g_\varepsilon$ satisfies \eqref{eq:defense2}, then $g_\varepsilon$ is an adversarial denoiser for $f$ on $S$.
\end{corollary}
\begin{proof}
    If $g_\varepsilon$ is a $\tau$-bounded denoiser for a given $\varepsilon$, then $\norm{g_\varepsilon(v) - v}_2 \leq \tau$ for all $v \in \mathcal V$. Because $f$ is $\tau$-robust, it follows that $(f \circ g_\varepsilon)(v) = f(v)$ for all $v \in \mathcal V$.
\end{proof}
Therefore, an adversarial denoiser can be obtained from a bounded denoiser that additionally has a performance guarantee on the reconstruction error. In the next section, we will show how to construct an adversarial denoiser based on the RWP, and how to use it to improve the robustness of a given classifier.

\subsection{Construction of a RWP-based adversarial defense}
\label{sec:RWP_defense_construct}

To take advantage of the RWP reconstruction performance guarantee from \cref{thm:rwp}, we can verify if the CS-RWP denoiser $\Delta_\varepsilon$ from \eqref{eq:delta} is an \textit{adversarial} denoiser for a $\tau$-robust classifier $f$ on $S \subset \mathcal V$:
\begin{equation}
    g_\varepsilon = \Phi \circ \Delta_\varepsilon.
\end{equation}
Here, $\Phi$ is a measurement operator satisfying the RWP. In the following, we will assume that $\Phi\Phi^\star = \ident_{\mathcal V}$, where $\Phi^\star$ is the adjoint of $\Phi$ and $\ident_{\mathcal V}$ is the identity operator on $V$. This property is satisfied by random tight frames (see, e.g., \cite{rauhut2019low}) and in particular the random partial Fourier matrices that we will use in \cref{sec:certification}. Our results can be extended to the non-tight frame case, but the performance of the adversarial defense is then less favorable.

\begin{lemma}
\label{lem:cs_bounded_denoiser}
    Let $(\mathcal V, \mathcal{A}, \norm{\cdot}_\sharp)$ be a CS space with bound $L$ and let $\Phi: \mathcal V' \to \mathcal V$ be a linear operator satisfying the $(\rho, \alpha)$-RWP with $\Phi\Phi^\star = \ident_{\mathcal V}$, for some $\rho < 1/(4L)$. Let $\lambda \geq 0$, then the denoiser $\Delta_\varepsilon$ satisfies:
    \begin{equation}
        \norm{\Phi(\Delta_\varepsilon(\Phi x' + \lambda e) - x')}_2 \leq \frac{(1+\lambda)\varepsilon}{\alpha} + 4\rho\defect(x')
    \end{equation}
    for all $x' \in \mathcal V$ and $e \in \mathcal V'$ with $\norm{e}_2 \leq \varepsilon$.
\end{lemma}
\begin{proof}
    Using $\Phi\Phi^\star = \ident_{\mathcal V}$ and given an input $x + \delta \in \mathcal V$, we can transform it into the domain of the sensing operator using $\Phi^\star(x + \delta) = \Phi^\star x + \Phi^\star\delta$. Letting $x' = \Phi^\star x$ and $e = \Phi\Phi^\star\delta$, we can then apply \cref{thm:rwp} to obtain
$$
    \norm{\Delta_\varepsilon(\Phi x' + e) - x'}_2 \leq \frac{2\varepsilon}{\alpha} + 4\rho\defect(x')
$$
provided $\norm{e}_2 = \norm{\delta}_2 \leq \varepsilon$. With straightforward adaptations of the proof of \cite[Theorem 3]{cahill2021robust}, this equation can be generalized to:
$$
    \norm{\Delta_\varepsilon(\Phi x' + \lambda e) - x'}_2 \leq \frac{(1 + \lambda)\varepsilon}{\alpha} + 4\rho\defect(x')
$$
for $\lambda \geq 0$. Since $\Phi\Phi^\star = \ident_{\mathcal V}$ we have
$$
    \norm{\Phi(\Delta_\varepsilon(\Phi x' + \lambda e) - x')}_2 \leq \norm{\Delta_\varepsilon(\Phi x' + \lambda e) - x'}_2.
$$
As a result,
\begin{align*}
    \norm{\Phi(\Delta_\varepsilon(\Phi x' + \lambda e) - x')}_2
    &= \norm{\Phi\Delta_\varepsilon(\Phi x' + \lambda e) - \Phi x'}_2\\
    &= \norm{\Phi\Delta_\varepsilon(x + \lambda \delta) - x}_2\\
    &\leq \frac{(1+\lambda)\varepsilon}{\alpha} + 4\rho\defect(\Phi^\star x), \\
\end{align*}
which concludes the proof.
\end{proof}

We now state our main result:

\begin{theorem}\label{prop:rwp}
Let $f: \mathcal V \to \mathcal W$ be a $\tau$-robust classifier in $S \subset \mathcal V$, where $\tau$ is a tight robustness bound. Let $(\mathcal V, \mathcal{A}, \norm{\cdot}_\sharp)$ be a CS space with bound $L$ and let $\Phi: \mathcal V' \to \mathcal V$ be a linear operator satisfying $\Phi\Phi^\star = \ident_{\mathcal V}$ and the $(\rho, \alpha)$-RWP, for some  $\rho \leq \frac{1}{4L}$. Then $\Delta_\varepsilon$ satisfies $\norm{\Phi(\Delta_\varepsilon(\Phi x' + e) - x')}_2 \leq C(\varepsilon) = \varepsilon \kappa(\varepsilon)$ for all $x' \in \mathcal V$ and $e \in \mathcal V'$, with 
\begin{equation}
\label{eq:k_epsilon}
    \kappa(\varepsilon) = \frac{2}{\alpha} + \frac{4\rho}{\varepsilon}\max_{x \in S}~\defect(\Phi^\star x).
\end{equation}
In addition, the following two statements hold:
\begin{enumerate}[label=(\alph*)]
    \item if either $\varepsilon \leq \tau$ or $C(\varepsilon) \leq \tau + \frac{\varepsilon}{\alpha}$, then the denoiser $g_\epsilon = \Phi \circ \Delta_\varepsilon$ is an adversarial denoiser for $f$ on $S$ with performance bound $C(\varepsilon)$, and
    \item if $C(\varepsilon) \leq \tau$, the smoothed classifier $f \circ g_\varepsilon = f \circ \Phi \circ \Delta_\varepsilon$ is $\varepsilon$-robust in $S$, with robustness gain of at least $1/\kappa(\varepsilon)$.
\end{enumerate}
\end{theorem}

\begin{proof}
    Let $x \in \mathcal V$, $x' = \Phi^\star x$ and $e = \Phi\Phi^\star\delta$. From \cref{lem:cs_bounded_denoiser} with respectively $\lambda=0$ and $\lambda=1$, it follows that the denoiser $\Delta_\varepsilon$ is a $\left(\varepsilon \kappa(\varepsilon) - \frac{\varepsilon}{\alpha}\right)$-bounded denoiser with performance bound $C(\varepsilon) = \varepsilon \kappa(\varepsilon)$, i.e.,
\begin{align*}
    \norm{\Phi(\Delta_\varepsilon(\Phi x' + e) - x')}_2
    &\leq \frac{2\varepsilon}{\alpha} + 4\rho\inf_{a \in \mathcal{A}}\norm{\Phi^\star x - a}_\sharp \\
    & = \varepsilon \left( \frac{2}{\alpha} + \frac{4\rho}{\varepsilon}\inf_{a \in \mathcal{A}}\norm{\Phi^\star x - a}_\sharp \right) \\
\end{align*}
Consequently, for all $x' \in \mathcal V'$ and $e \in \mathcal V$ with $\norm{e}_2 \leq \varepsilon$,
$$
\norm{\Phi(\Delta_\varepsilon(\Phi x' + e) - x')}_2 \leq C(\varepsilon) = \varepsilon \kappa(\varepsilon).
$$
with $\kappa(\varepsilon) =  \frac{2}{\alpha} + \frac{4\rho}{\varepsilon}\max_{x \in S} \inf_{a \in \mathcal{A}}\norm{\Phi^\star x - a}_\sharp$. 
\begin{enumerate}
    \item To prove (a):  $\varepsilon \leq \tau$ or $C(\varepsilon) = \varepsilon \kappa(\varepsilon) \leq \tau + \frac{\varepsilon}{\alpha}$ imply that $(f \circ g_\epsilon)(v) = f(v)$ for all $v \in \mathcal V$, by \cref{def:sample_robustness} and \cref{cor:bounded_denoiser}. Therefore, according to \cref{def:adv_denoiser}, the denoiser $\Delta_\varepsilon$ is an adversarial denoiser for $f$ on $S$ with performance bound $C(\varepsilon)$.
    \item
    To prove (b): because $f$ is $\tau$-robust and $C(\varepsilon) \leq \tau$, this implies $f(\Phi(\Delta_\varepsilon(\Phi x' + e))) = f(\Phi x') = f(x)$, i.e., $f \circ g_\epsilon$ is $\epsilon$-robust in $S$. In addition, applying \cref{prop:adv_denoiser_robustness} yields the robustness gain of at least $1/\kappa(\varepsilon)$.
\end{enumerate} 
\end{proof}

\Cref{prop:rwp} shows that the denoiser $\Delta_\varepsilon$ improves the robustness bound $\tau$ of a given classifier $f$ on a sample set $S$, provided that $C(\varepsilon) < \tau$. In this case, the performance bound is given by
$$
    C(\varepsilon) = \frac{2\varepsilon}{\alpha} + 4\rho\defect(\Phi^\star x).
$$
Hence, the condition $C(\varepsilon) < \tau$ can be translated into a sparsity requirement of the data:
\begin{equation}
\label{eq:sparsity_defect_bound}
    \max_{x \in S}~\defect(\Phi^\star x) < \frac{1}{4\rho}\left( \tau - \frac{2\varepsilon}{\alpha} \right).
\end{equation}
This condition is only meaningful if $\tau \alpha/2 > \varepsilon$. In other words, to correctly assess the robustness improvement, it is preferable that $\tau$ is a tight robustness bound. Moreover, the sparsity requirement \eqref{eq:sparsity_defect_bound} becomes less stringent as $\rho$ becomes smaller and $\alpha$ becomes larger. Therefore, all else being equal, an operator $\Phi$ satisfying the $(\rho,\alpha)$-RWP will lead to a better adversarial defense than an operator $\Phi'$ satisfying the $(\rho',\alpha')$-RWP with $\rho < \rho'$ and $\alpha > \alpha'$.

\subsection{Probabilistic adversarial defense method}

The practical utility of \cref{prop:rwp} is limited by the fact that, in practice, the sensing operator $\Phi$ is not deterministic but rather sampled from some random matrix ensemble, due to the availability of a RWP proof for this setting. The construction of deterministic sensing operators which also admit efficient implementations is a major avenue of ongoing research~\citep{monajemi2013deterministic,yu2013deterministic}. Therefore, we will only be able to guarantee that the operator satisfies the RWP with some high probability. This leads to a probabilistic version of \cref{prop:rwp}:

\begin{theorem}\label{thm:adv}
Let $f: \mathcal V \to \mathcal W$ be a classifier and let $x \in S \subset \mathcal V, \delta \in \mathcal V$ with $\norm{\delta}_2 \leq \varepsilon$. Assume $\Phi\Phi^\star = \ident_{\mathcal V}$ and let $f$ be $\tau$-robust in $S$, with $\varepsilon \kappa(\varepsilon) \leq \tau$ and $\kappa(\varepsilon)$ as in \eqref{eq:k_epsilon}. Let $\Phi$ satisfy the $(\rho,\alpha)$-RWP with probability at least $q$. Then we will have $f(\Phi\Delta_\varepsilon(x + \delta)) = f(x)$ for all $x \in S$ with probability at least
$$
    q - \frac{4\alpha \rho}{\alpha\tau - 2\varepsilon}\expect\left[ \max_{x \in S}~\defect(\Phi^\star x) \right],
$$
where the expectation is taken over $\Phi$.
\end{theorem}
\begin{proof}
For brevity, we denote
$$
    \defect = \max_{x \in S}~\defect(\Phi^\star x).
$$
By \cref{prop:rwp}, the probability that $f(\Phi\Delta_\varepsilon(x + \delta)) = f(x)$ for every $x \in S$ is lower bounded by the joint probability that $\Phi$ satisfies the RWP and the sparsity defect is sufficiently small (see \eqref{eq:sparsity_defect_bound}):
\begin{align*} 
    \Pr[\forall x \in S: f(\Phi\Delta_\varepsilon(x + \delta)) = f(x)]
    &\geq \Pr\left[\mbox{$\Phi$ satisfies the RWP} \wedge \defect < \frac{1}{4\rho}\left( \tau - \frac{2\varepsilon}{\alpha} \right) \right]\\
    &\geq \Pr[\mbox{$\Phi$ satisfies the RWP}] + \Pr\left[\defect < \frac{1}{4\rho}\left( \tau - \frac{2\varepsilon}{\alpha} \right) \right] - 1\\
    &\geq q + \Pr\left[\defect < \frac{1}{4\rho}\left( \tau - \frac{2\varepsilon}{\alpha} \right) \right] - 1
\end{align*}
where the second inequality follows from the intersection bound. We then apply Markov's inequality:
$$
    \Pr\left[\defect < \frac{1}{4\rho}\left( \tau - \frac{2\varepsilon}{\alpha} \right) \right]
    \geq 1 - \frac{4\alpha\rho}{\alpha\tau - 2\varepsilon}\expect\left[ \defect \right].
$$
This yields the stated result.
\end{proof}

The robustness bound guaranteed by \cref{thm:adv} is directly controlled by parameters that are easily interpretable:
\begin{itemize}
    \item The RWP parameters $\alpha$ and $\rho$ of the sensing operator $\Phi$, where preferably $\alpha$ is large and $\rho$ is small. According to the RWP, the null space of $\Phi$ intersects the unit $\sharp$-ball $B_\sharp$ with a maximal radius $\leq \rho$. Then every point of $\mathcal A$ should be sufficiently far away from this null space, resulting in a large value for $\alpha$.

    \item The expected sparsity defect of the data samples, which quantifies the extent to which the space $\mathcal A$ accurately captures the data manifold as well as the effect of random noise. In particular, if the data samples are \textit{exactly} sparse with respect to the chosen frame, then robustness holds with probability at least $q$.
\end{itemize}
Note that we can only certify robustness for $\varepsilon < \tau\alpha/2$, so we must have $\alpha > 2$ and the classifier must meet a minimum robustness requirement. A classifier that does not meet this minimum robustness cannot benefit from our defense. Although this may seem like a drawback, existing certified methods must also make this assumption. In the case of RS, for example, a tacit robustness assumption on the base classifier is made in the form of the probability margin $\Phi^{-1}(p_1) - \Phi^{-1}(p_2)$, which can be turned into a robustness requirement via continuity properties of $f$. In general, it is not possible to increase robustness of a classifier that does not already possess some minimum robustness without sacrificing accuracy. As a contrived example, consider the Dirichlet function:
$$
    f(x) = \begin{cases}
        1 & \mbox{if $x \in \mathbb Q$,}\\
        0 & \mbox{otherwise.}
    \end{cases}
$$
The rationals and irrationals are both dense in $\real$, so this function has zero robustness at any $x \in \real$. Applying RS to $f$ will yield a constant smoothed classifier $\hat{f}(x) = 0$, since $x + \xi$ will almost surely be irrational for any $\xi \sim \mathcal N(0, \sigma^2)$. If the data distribution has an \textit{a priori} probability $p$ that any data sample is rational, then the smoothed classifier $\hat{f}$ has a standard accuracy of $1 - p$.

\Cref{thm:adv} allows us to certify the adversarial defense $\hat{f} = f \circ \Phi \circ \Delta_\varepsilon$ probabilistically when $\Phi$ is sampled from a random matrix ensemble. To select $\Phi$, one particular class of commonly used sensing operators which satisfy the requirements of \cref{thm:adv}, are random Fourier matrices. Specifically, if $\Phi$ is a $\mathbb C^{m \times N}$ matrix constructed by taking $m < N$ rows uniformly at random from an $N \times N$ Fourier matrix, then $\Phi$ always satisfies $\Phi\Phi^\star = I_m$ and will satisfy the RWP with high probability. More precisely, random Fourier matrices are known to satisfy the so-called \textit{restricted isometry property} (RIP) with high probability~\citep{haviv2017restricted}. As shown by \citet{cahill2021robust}, the RIP implies the RWP and so partial Fourier matrices satisfy the RWP needed for \cref{thm:rwp} with high probability. In contrast to RIP, RWP additionally gives recovery guarantees for \textit{approximately} sparse images.  Aside from fulfilling the theoretical requirement, Fourier matrices are also computationally efficient because the matrix multiplication $\Phi x$ can be implemented without evaluating $\Phi$ explicitly. Indeed, to determine $\Phi x$, we can simply compute the Fourier transform of $x$, zero out a random subset of the coefficients, and then apply the inverse Fourier transform. Or, in a more advanced implementation the erased coefficients do not even need to be calculated at all (see e.g., \cite{desmet2004}).

\begin{algorithm}
    \begin{algorithmic}
        \Procedure {Reconstruct}{input $x \in \mathbb R^n$, transform $\Psi: \mathbb R^n \to \mathbb R^n$, number of iterations $T \in \mathbb N$, threshold $\lambda \in \mathbb R$, subsampling probability $q \in [0,1]$}
            \State Sample a binary mask $m \sim \mathrm{Ber}(q)^n$.
            \State Define
            \begin{align*}
                \Phi(x) &= m \odot \mathrm{FFT}(x), & \Phi^\star(y) &= \mathrm{IFFT}(m \odot y).
            \end{align*}
            \State $u_0 \gets 0$
            \State $t \gets 1$
            \While{$t \leq T$}
                \State $u_t \gets \mathcal{S}_\lambda\left( u_{t-1} + \Psi\Phi^\star\left( y - \Phi\Psi^{-1} u_{t-1} \right) \right)$
                \State $t \gets t + 1$
            \EndWhile
            \State Return $\Psi^{-1} u_T$
        \EndProcedure
    \end{algorithmic}
    \caption{Robust width adversarial defense.}
    \label{alg:rwp}
\end{algorithm}

The pseudo-code of our defense is given in \cref{alg:rwp}. Here, $\mathcal{S}_\lambda$ is the soft-thresholding function:
\[
    \mathcal{S}_\lambda(u) = \begin{cases}
        0 & \mbox{if $|u| < \lambda$,}\\[2mm]
        \displaystyle\frac{u}{|u|}(|u| - \lambda) & \mbox{otherwise.}
    \end{cases}
\]
This function is applied component-wise to vector-valued inputs. For real-valued inputs, this function simplifies to
\[
    \mathcal{S}_\lambda(u) = \mathrm{sign}(x) \cdot \max\{ 0, |x| - \lambda \}.
\]
We use the more general formula above because the coefficients $u$ can be complex-valued, e.g., when $\Psi$ is the Fourier or dual-tree complex wavelet transform.

The robustness guaranteed by \cref{thm:adv} applies to a single realization of the sensing operator $\Phi$. This result may therefore be improved using, \textit{e.g.}, an ensemble averaging or majority voting scheme across multiple samples of $\Phi$, similar to the multiple noise samples of randomized smoothing. Moreover, the dependency on the expected sparsity defect of the unperturbed data samples may be improved by only retaining samples of $\Phi$ that maximize sparsity. Such improvements are an interesting avenue for future work. Finally, we note that our defense enjoys a certain \textit{blessing} of dimensionality: as the data dimensionality increases, the probability $q$ that the random Fourier matrix will satisfy the RWP tends to one \textit{exponentially fast} regardless of $\alpha$ and $\rho$~\citep{haviv2017restricted}. By contrast, RS can suffer from a rather severe curse of dimensionality where the certifiable radius decays rapidly in the input dimension~\citep{sukenik2021intriguing}. In our case, as long as the data are approximately sparse, the dimensionality does not matter.

\subsection{Certification}
\label{sec:certification}
\Cref{thm:adv} shows that the robustness of our method can be probabilistically certified. To make practical use of this result, however, a method is needed for computing the bound. This is not straightforward, as it requires knowledge of several variables: $q$, the probability that the sensing operator satisfies the RWP; the parameters $\alpha$ and $\rho$ of the RWP; the robustness $\tau$ of the base classifier and the expected sparsity defect.

The probability that our sensing operator satisfies the RWP can be derived by combining the results of \cite[theorem 4.5]{haviv2017restricted} and \cite[theorem 11]{cahill2021robust}.

\begin{corollary}
A random partial Fourier matrix $A \in \mathbb C^{m \times N}$ where
$$
    m = O\left( \frac{1}{(1/3 - \alpha)^2}\log^2\frac{1}{1/3 - \alpha} \cdot \rho^{-2} \cdot \log^2\frac{9}{\rho^2(1/3 - \alpha)} \cdot \log N \right)
$$
satisfies the $(\rho,\alpha)$-RWP with probability $1 - 2^{-\Omega\left(\log N \cdot \left( \log 9 - \log(\rho^2(1/3 - \alpha)) \right) \right)}$ provided $\alpha < 1/3$.
\end{corollary}
\begin{proof}
From \cite[theorem 4.5]{haviv2017restricted} we have that $A$ satisfies the $(J, \delta)$-RIP with probability $1 - 2^{-\Omega(\log N \cdot \log(J/\delta))}$. \cite[theorem 11]{cahill2021robust} then shows $A$ also satisfies the $(\rho,\alpha)$-RWP where
\begin{align*}
    \rho &= \frac{3}{\sqrt{J}}, & \alpha &= \frac 13 - \delta
\end{align*}
with probability $1 - 2^{-\Omega(\log N \cdot \log(J/\delta))}$ provided $\delta < 1/3$. It follows that
\begin{align*}
    J &= \frac{9}{\rho^2}, & \delta &= \frac 13 - \alpha.
\end{align*}
Hence $A$ satisfies the $(\rho,\alpha)$-RWP with probability
$$
    1 - 2^{-\Omega(\log N \cdot \log(J/\delta)} = 1 - 2^{-\Omega\left(\log N \cdot \left( \log 9 - \log(\rho^2(1/3 - \alpha)) \right) \right)}
$$
provided $\alpha < \frac 13$.
\end{proof}

To compute the robustness of the base classifier, we can use existing robustness verification algorithms~\citep{brix2023fourth,wang2021beta} or strong adversarial attacks which explicitly minimize perturbation magnitude~\citep{croce2020minimally,carlini2017towards}. 

The expected sparsity defect can be estimated based on a finite set of samples $\Phi_1, \dots, \Phi_s$ from the random matrix ensemble:
$$
    \expect\left[ \max_{x \in S}\inf_{a \in \mathcal{A}}\norm{\Phi^\star x - a}_\sharp \right]
    \approx \frac{1}{s}\sum_{i=1}^s\max_{x \in S}\inf_{a \in \mathcal{A}}\norm{\Psi(\Phi_i^\star x - a)}_1.
$$
For a solution space consisting of functions in $\mathcal H$ with $T$-bounded $\sharp$-norm,
$
    \mathcal A = \{ x \in \mathcal H \mid \norm{x}_\sharp \leq T \},
$
and a sensing operator $\Phi$, determining the infimum becomes a constrained $L^1$ optimization problem:
\begin{equation}\label{eq:cert1}
    \inf_{a \in \mathcal H}~\norm{\Psi(\Phi^\star x - a)}_1 \mbox{ such that } \norm{\Psi a}_1 \leq T.
\end{equation}
We can convert this into an unconstrained problem using the split-Bregman method~\citep{goldstein2009split}. Introducing the splitting variable $a' = \Psi a$, we can rewrite \eqref{eq:cert1} as
$$
    \inf_{a' \in \mathcal A'}~\norm{\Psi\Phi^\star x - a'}_1 \mbox{ such that } \norm{a'}_1 \leq T
$$
where $\mathcal A' = \Psi(\mathcal A)$. The solution is given by initializing $b_0 = 0$, $a_0' = \Psi\Phi^\star x$ and iterating
\begin{align}
    a_{i+1}' &= \arg\min_{a' \in \mathcal A'}~\norm{a'}_1 + \frac{\lambda}{2}\norm{a' - \Psi\Phi^\star x - b_i}_1,\label{eq:cert2}\\
    b_{i+1} &= b_i + \Psi\Phi^\star x - a_{i+1}',
\end{align}
until $\norm{a_i'}_1 \leq T$. The choice of $\lambda$ will not affect the correctness of the solution, but may impact its rate of convergence. The minimization \eqref{eq:cert2} is an $L^1+L^1$ problem, so we may further rewrite it as
\begin{align*}
    u_{i+1} &= \mathcal{S}_\lambda(d_i - b_i - \Psi\Phi^\star x) + \Psi\Phi^\star x,\\
    d_{i+1} &= \mathcal{S}_\lambda(u_{i+1} + b_i),\\
    b_{i+1} &= b_i + u_{i+1} - d_{i+1}.
\end{align*}
The procedure is summarized in \cref{alg:bound_l1_minimization}.

\begin{algorithm}
    \begin{algorithmic}
        \Procedure {CalculateBound}{$\Phi$, $\Psi$, $x$, $T$, $\lambda$, $\mathsf{tolerance}$}
        \State Initialize $d_0 = \Psi\Phi^\star x$, $b_0 = 0$, $d_{-1} = 0$, $i = 0$
        \While{$\norm{d_i - d_{i - 1}}_1 > \mathsf{tolerance}$}
            \State $u_{i + 1} \gets \mathcal{S}_\lambda\left(d_i - b_i - \Psi\Phi^\star x\right) + \Psi\Phi^\star x$
            \State $d_{i + 1} \gets \mathcal{S}_\lambda\left(u_{i+1} + b_i \right)$
            \State $b_{i + 1} \gets b_i + (u_{i+1} - d_{i + 1})$
            \State $i \gets i + 1$
        \EndWhile
        \If{$\norm{d_i}_1 \leq T$}
            \State The solution is $\norm{\Psi\Phi^\star x - d_i}_1$
        \Else
            \State No solution was found
        \EndIf
        \EndProcedure
    \end{algorithmic}
    \caption{Procedure to compute the sparsity defect of a given sample.}
    \label{alg:bound_l1_minimization}
\end{algorithm}

\subsection{Example: linear classification of sparse vectors}
To illustrate the main ideas of this work, we turn to the classic setting of sparse vectors as a simple example. Specifically, we let $\mathcal A$ be the set of $K$-sparse vectors for some $K > 0$, i.e., $\mathcal A$ consists of vectors in $\real^N$ with at most $K$ non-zero components. The sparsity norm $\norm{\cdot}_\sharp$ is given by
$$
    \norm{x}_\sharp = \norm{x}_1 = \sum_{i=1}^N|x_i|.
$$
Then we have a CS space with bound $L = \sqrt{K}$~\citep{cahill2021robust}. \Cref{thm:rwp} yields the reconstruction guarantee
$$
    \norm{\Delta_\varepsilon(\Phi x^\natural + e) - x^\natural}_2 \leq \frac{2\varepsilon}{\alpha} + 4\rho\defect(x^\natural)
$$
provided $\rho \leq \frac{1}{4\sqrt{K}}$ and $\Phi$ satisfies the $(\rho,\alpha)$-RWP. Assuming the signals $x^\natural$ are \textit{exactly} $K$-sparse, this bound reduces to
$$
    \norm{\Delta_\varepsilon(\Phi x^\natural + e) - x^\natural}_2 \leq \frac{2\varepsilon}{\alpha}.
$$
In other words, we have
$$
    \kappa(\varepsilon) = \frac 2\alpha.
$$
We define a binary classifier on $\real^N$ given by
$$
    f(x) = \mathrm{sgn}(\inprod{w}{x})
$$
where $w \in \real^N$ is a vector of weights. The robustness of such a classifier on a singleton set $S = \{ x \}$ is precisely
$$
    \tau = \frac{|\inprod{w}{x}|}{\norm{w}_2}.
$$
The condition $\varepsilon \kappa(\varepsilon) \leq \tau$ then becomes
$$
    \frac{2\varepsilon}{\alpha} \leq \frac{|\inprod{w}{x}|}{\norm{w}_2}
    \iff \varepsilon \leq \frac{\alpha|\inprod{w}{x}|}{2\norm{w}_2}.
$$
Furthermore, $\kappa(\varepsilon) < 1$ implies $\alpha > 2$. Putting this all together, we have the following result:

\begin{theorem}\label{thm:sparse_example}
Let $\Phi$ satisfy the $(\rho,\alpha)$-RWP with $\alpha > 2$. Define the linear classifier
$$
    f(x) = \mathrm{sgn}(\inprod{w}{x}).
$$
Then, given any $K$-sparse vector $x \in \real^N$, it holds that $f(\Phi\Delta_\varepsilon(x + \delta)) = f(x)$ for all perturbations $\delta$ with
$$
    \norm{\delta}_2 \leq \frac{\alpha|\inprod{w}{x}|}{2\norm{w}_2}.
$$
The robustness gain is at least $\alpha / 2$.
\end{theorem}

\Cref{thm:sparse_example} illustrates how large robustness gains may be achieved depending on the parameters of the RWP as well as the data samples themselves. For instance, we may choose $x$ such that $\inprod{w}{x} = 0$. These samples lie exactly on the decision boundary and so the classifier has zero robustness around them, leading to the trivial certificate $\norm{\delta}_2 \leq 0$. On the other hand, if $\inprod{w}{x} \neq 0$, our method provides a robustness gain of at least $\alpha/2$ provided $\alpha > 2$. Large values of $\alpha$ in this case directly correspond to large gains in robustness.

The parameter $\rho$ does not appear in \cref{thm:sparse_example}, because we have assumed exactly sparse signal vectors and $\rho$ only plays a role in controlling the sparsity defect. If the data are not exactly $K$-sparse, then we have instead
$$
    \kappa(\varepsilon) = \frac 2\alpha + \frac{4\rho}{\varepsilon}\defect(\Phi^\star x).
$$
The condition $\varepsilon \kappa(\varepsilon) \leq \tau$ then becomes
$$
    \frac{2\varepsilon}\alpha + 4\rho\defect(\Phi^\star x) \leq \frac{|\inprod{w}{x}|}{\norm{w}_2}
    \iff \varepsilon \leq \frac\alpha 2\left( \frac{|\inprod{w}{x}|}{\norm{w}_2} - 4\rho\defect(\Phi^\star x) \right).
$$
This leads to
\begin{theorem}\label{thm:sparse_example_2}
Let $\Phi$ satisfy the $(\rho,\alpha)$-RWP with $\alpha > 2$. Define the linear classifier
$$
    f(x) = \mathrm{sgn}(\inprod{w}{x}).
$$
Let $x \in \real^N$ such that
$$
    \frac{|\inprod{w}{x}|}{\norm{w}_2} > 4\rho\defect(\Phi^\star x).
$$
Then $f(\Phi\Delta_\varepsilon(x + \delta)) = f(x)$ for all perturbations $\delta$ with
$$
    \norm{\delta}_2 \leq \frac\alpha 2\left( \frac{|\inprod{w}{x}|}{\norm{w}_2} - 4\rho\defect(\Phi^\star x) \right).
$$
The robustness gain is at least
$$
    \frac{1}{\kappa(\varepsilon)} = \left( \frac 2\alpha + \frac{4\rho}{\varepsilon}\defect(\Phi^\star x) \right)^{-1}.
$$
\end{theorem}

We see that \cref{thm:sparse_example_2} requires some robustness of the underlying classifier before it can guarantee improvement. Because the classifier is linear, this robustness requirement takes the form of a minimal distance from the hyperplane spanned by $w$. Equivalently, it can be interpreted as a minimum sparsity requirement imposed on the data:
$$
    \frac{|\inprod{w}{x}|}{\norm{w}_2} > 4\rho\defect(\Phi^\star x)
    \iff
    \defect(\Phi^\star x) < \frac{|\inprod{w}{x}|}{4\rho\norm{w}_2}.
$$
For $K$-sparse vectors, this bound becomes vacuous and the result essentially reduces to \cref{thm:sparse_example}; for vectors that are not $K$-sparse, the bound increases linearly with $\rho$, so smaller values of $\rho$ imply better tolerance for sparsity defects and fragility of the underlying classifier.

Note, however, that \cref{thm:sparse_example_2} is overly pessimistic in many cases: in practice, linear classifiers are often trained using a regularized loss function which enforces sparsity of the weight vector $w$~\citep{hastie2009elements}. If we can assume $w$ itself is $K$-sparse, then $f(x) = f(m \odot x)$ for all $x \in \real^N$, where $\odot$ denotes element-wise multiplication and $m$ is a mask defined by
$$
    m_i = \begin{cases}
        1 & \mbox{if $w_i \neq 0$,}\\
        0 & \mbox{otherwise.}
    \end{cases}
$$
The sparsity defect of the data then becomes functionally zero, so we can discard it and apply \cref{thm:sparse_example} instead. This provides a very direct justification for weight sparsity which may be generalized to more complicated settings, such as deep neural networks.  

\section{Experiments}
\label{sec:experiments}

To summarize, our adversarial defense takes the form of a denoiser $g_\varepsilon = \Phi \circ \Delta_\varepsilon$ which is prepended to the classifier. Here, $\Delta_\varepsilon$ is defined as in \eqref{eq:delta} and $\Phi$ is a random partial Fourier matrix. For the sparsity norm $\norm{\cdot}_\sharp$, we restrict ourselves to $L^1$ norms in an appropriate vector space:
$$
    \norm{u}_\sharp = \sum_{j=1}^n|\inprod{u}{\psi_j}|
$$
where $\{\psi_j\}_{j=1}^n$ is a frame of $\real^n$. We experiment with several possibilities for the choice of frame, including various wavelet bases~\citep{graps1995introduction}, the Fourier basis and shearlet bases~\citep{guo2006sparse}. Pseudocode of our method is provided in \cref{alg:rwp}.

Our experiments are carried out on the ImageNet data set~\citep{deng2009imagenet} using pre-trained classifiers implemented in PyTorch~\citep{pytorch}, including WideResNet-101~\citep{zagoruyko2016wide}, ResNet-50~\citep{he2016deep}, Vision Transformer~\citep{dosovitskiy2020image} and Swin Transformer~\citep{liu2021swin}. The implementation of the wavelet and shearlet transforms were provided by Pytorch Wavelets~\citep{pytorch_wavelets} and pyShearLab~\citep{pyshearlab}, respectively. All adversarial attacks are implemented by the AutoAttack library~\citep{croce2020reliable} and the Adversarial Robustness Toolbox~\citep{art2018}. We compare our robust accuracy scores to several published models from the RobustBench library~\citep{croce2020robustbench}: the adversarially trained Swin Transformer by \citet{liu2023comprehensive}, the XCiT-L12 model by \citet{debenedetti2023light}, the RaWideResNet introduced by \citet{peng2023robust} and the adversarially trained ResNet-50 by \citet{wong2020fast}. In the white-box setting, we also compare against the robustness of DiffPure~\citep{nie2022diffusion}.

\subsection{Threat models}
We evaluate the models in the following settings:
\begin{itemize}
    \item \textbf{Black-box, no surrogate.} In this threat model, the adversary is assumed to have no direct access to the target model nor do they have the means to perform transfer attacks based on surrogate models. These experiments are carried out by directly attacking the models using the Square attack~\citep{andriushchenko2020square}, which is a powerful black-box attack that assumes only query access to the target.

    \item \textbf{Black-box, surrogate.} Here, the adversary still does not have direct access to the target model, but they are capable of obtaining a surrogate which can be used to launch a transfer attack~\citep{papernot2017practical}. These experiments are carried out by attacking a given surrogate model using APGD~\citep{croce2020reliable} and then evaluating the target models on these adversarial examples.
    
    \item \textbf{White-box.} In this setting, we directly attack our defended models using the randomized version of AutoAttack~\citep{croce2020reliable}. The randomized version is necessary due to the stochastic sensing matrix $\Phi$. Since the soft-thresholding function $\mathcal S_\lambda$ used in our denoiser $g_\varepsilon = \Phi \circ \Delta_\varepsilon$ is not differentiable everywhere, there may be gradient masking issues~\citep{athalye2018obfuscated}. We perform additional experiments to assess this potential problem according to the recommendations set forth by \citet{croce2022evaluating}.
\end{itemize}
All attacks are carried out in the $L^\infty$ norm with perturbation budgets ranging from 4/255 to 32/255.

\subsection{Hyperparameter tuning}
\begin{figure}[p]
    \begin{center}
        \subcaptionbox{WideResNet-101}{\includegraphics[width=.45\textwidth]{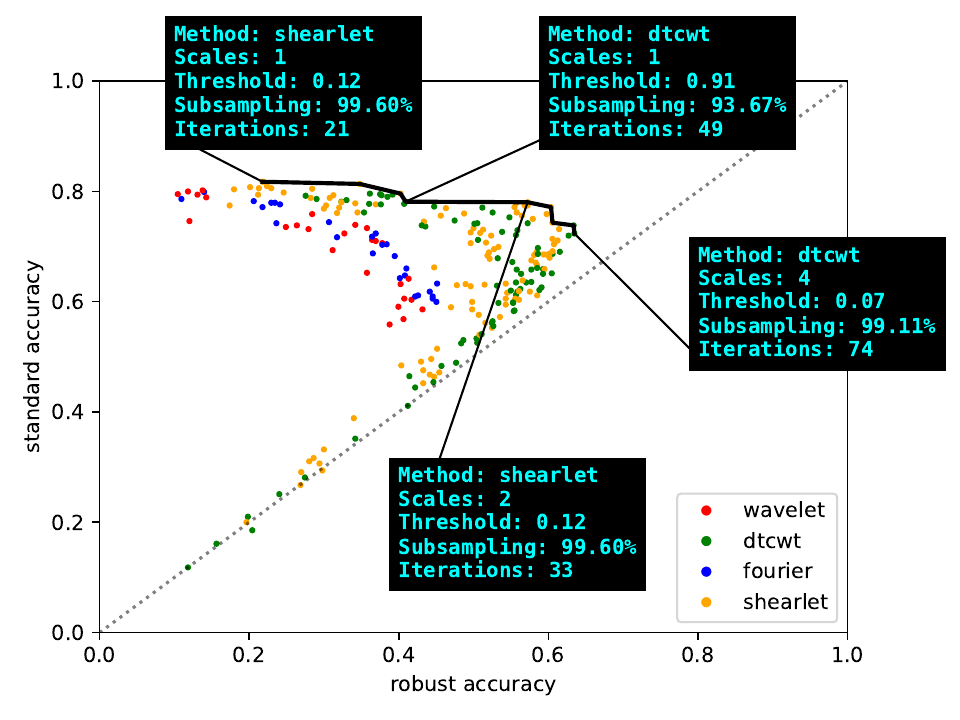}}
        \subcaptionbox{ResNet-50}{\includegraphics[width=.45\textwidth]{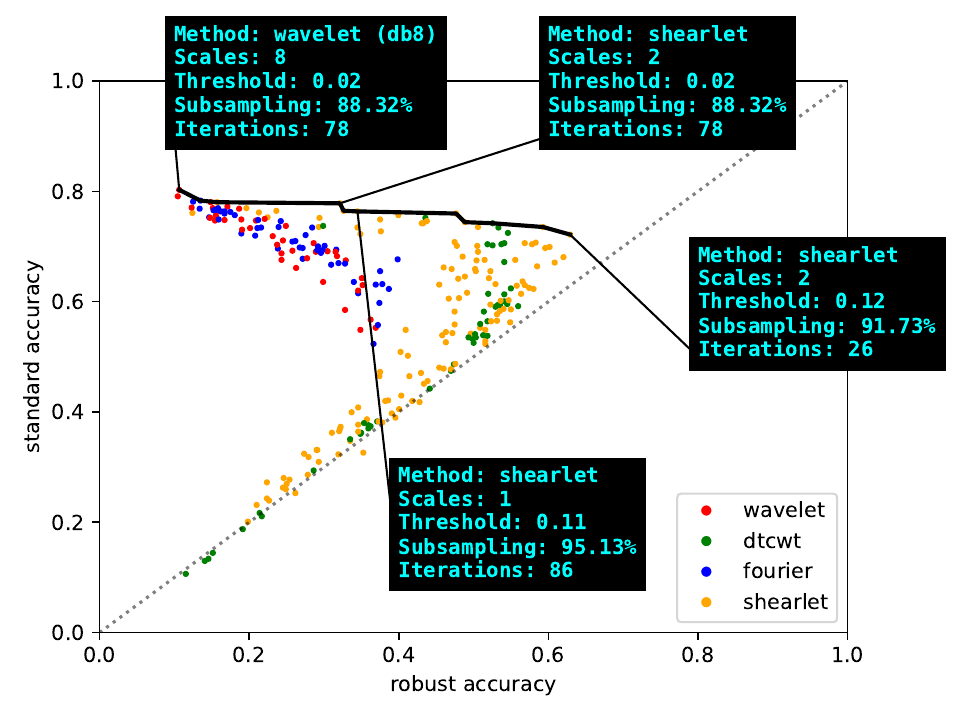}}
        
        \subcaptionbox{ViT-B-16}{\includegraphics[width=.45\textwidth]{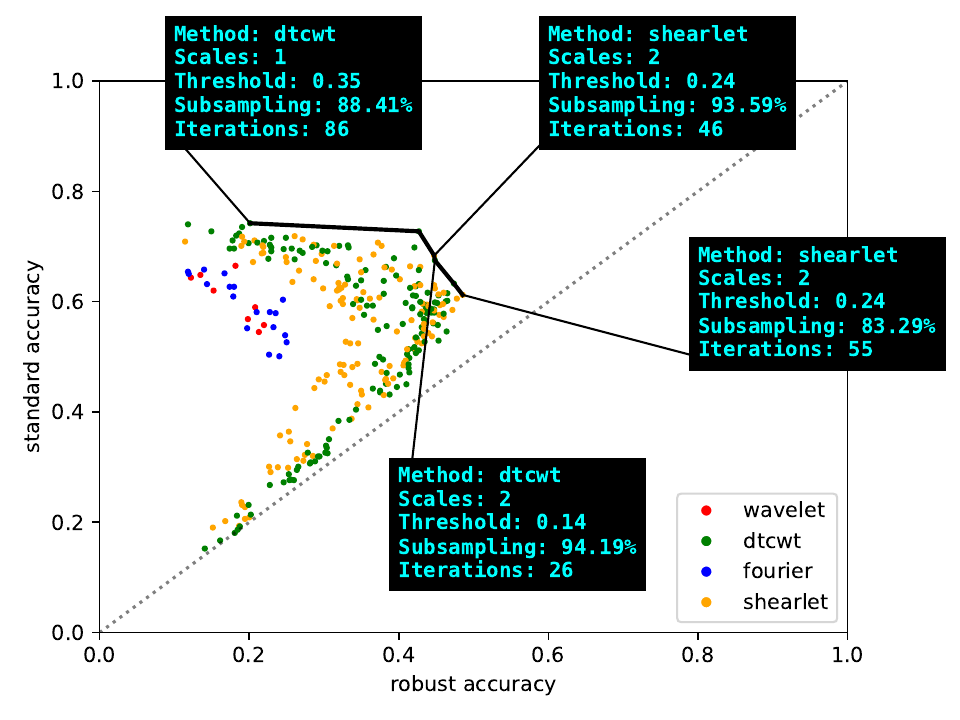}}
        \subcaptionbox{Swin-T}{\includegraphics[width=.45\textwidth]{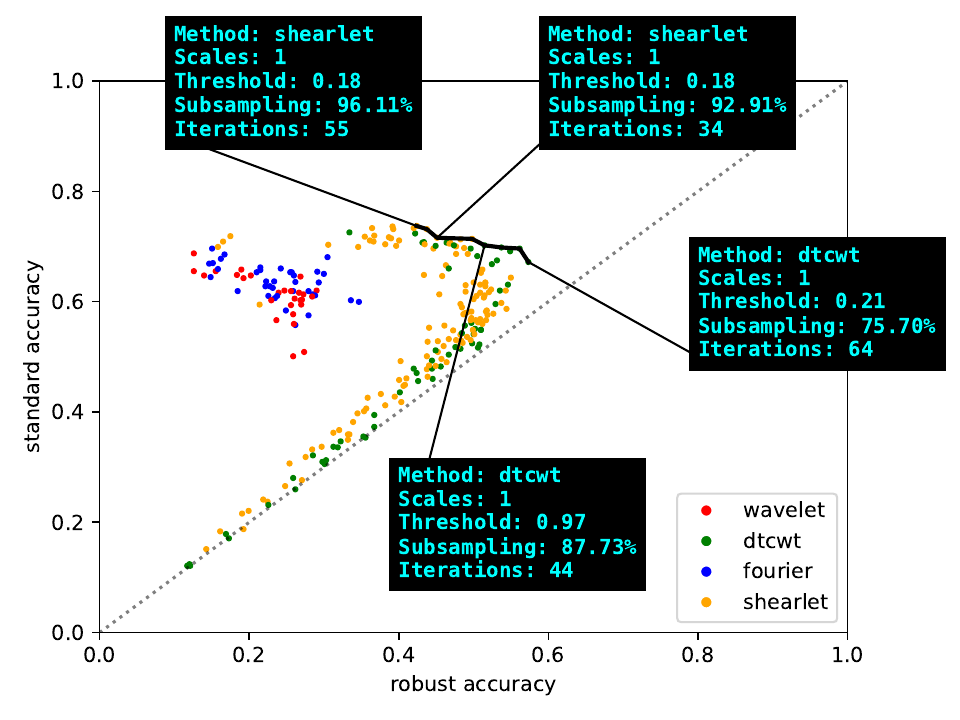}}
    \end{center}
    \caption{Hyperparameter search results for the different ImageNet classifiers. Trials which obtained less than 10\% standard or robust accuracy have been omitted. The Pareto front is highlighted in black, and some hyperparameter configurations along the front are given for illustration.}
    \label{fig:optuna}
\end{figure}

The hyperparameters of the defense were tuned using Optuna~\citep{akiba2019optuna} with 1,000 trials on a held-out validation set of 1,024 ImageNet samples. Given a configuration of hyperparameters selected by Optuna, we evaluate the robustness of the defense in the black-box, no surrogate setting, \ie we use $L^\infty$ APGD~\citep{croce2020reliable} with a budget of $4/255$ to adversarially attack the base classifier and then evaluate the smoothed model on these samples. \Cref{fig:optuna} shows the results of our hyperparameter tuning for the different classifiers. We plot the robust accuracy on the horizontal axis and the standard accuracy on the vertical axis. The results are color-coded according to the sparsity basis with which they were obtained (wavelet, DTCWT, Fourier or shearlet). The Pareto front is highlighted in black, and several hyperparameter configurations are given for illustration purposes. Based on these results, we selected the hyperparameter settings given in \cref{tab:parameters} to perform the robustness evaluations. The meaning of the different hyperparameters is explained in \cref{tab:hyper}.

\begin{table}[H]
    \centering
    \begin{tabular}{ll}
        \toprule
        Basis & Basis $\Psi$ chosen for sparsification of the input data\\
        Levels & Number of levels of decomposition in the chosen basis\\
        Threshold & Threshold value used in the iterative thresholding algorithm\\
        Iterations & Number of iterations of thresholding used for denoising\\
        Subsampling & Fraction of rows retained in the random partial Fourier matrix $\Phi$\\
        \bottomrule
    \end{tabular}
    \caption{Overview of the hyperparameters of our defense algorithm.}
    \label{tab:hyper}
\end{table}

\begin{table}[H]
    \centering
    \begin{tabular}{llrrrr}
        \toprule
         Model & Basis & Levels & Threshold & Iterations & Subsampling\\
         \midrule
         ResNet-50 & Shearlet & 2 & 0.12 & 26 & 91.73\%\\ 
         WideResNet-101 & DTCWT & 2 & 0.07 & 51 & 87.94\%\\ 
         ViT-B-16 & Shearlet & 2 & 0.24 & 55 & 83.28\%\\ 
         Swin-T & DTCWT & 1 & 0.21 & 64 & 75.70\%\\ 
         \bottomrule
    \end{tabular}
    \caption{Selected hyperparameter settings of the defense for each model.}
    \label{tab:parameters}
\end{table}

\subsection{Results}

\paragraph{Black-box, no surrogate.} The results in the black-box, no surrogate setting are reported in \cref{tab:square}. The first four entries report accuracy scores obtained by prepending our defense to pre-trained classifiers; the other entries compare to existing defenses taken from the RobustBench leaderboard.\footnote{\url{https://robustbench.github.io/\#div_imagenet_Linf_heading}. Accessed 2024-05-25.} We find that the standard accuracy of our defended models is slightly lower than the standard accuracy of existing robust models. For small perturbation budgets ($\varepsilon \leq 8/255$), there is no significant difference between our defense and the state of the art. However, for large perturbation budgets ($\varepsilon > 8/255$), the adversarially trained models suffer a rapid decline in robust accuracy while our defense does not: we maintain high robust accuracy even at $\varepsilon = 32/255$. Among the models we tested, the best performing one seems to be the WideResNet by a substantial margin.

\begin{table}[h]
    \begin{center}\footnotesize
        \begin{tabular}{lrrrr}
            \toprule
                    & & \multicolumn{3}{c}{Robust}\\
                    \cmidrule(lr){3-5}
             Model & Standard & $\varepsilon = 8/255$ & $\varepsilon = 16/255$ & $\varepsilon = 32/255$\\
             \midrule
             WRN-101-2 & 72.80\% $\pm$ 2.76\% & 58.00\% $\pm$ 3.06\% & \textbf{57.10\%} $\pm$ 3.07\% & \textbf{51.10\%} $\pm$ 3.10\%\\
             ResNet50  & 68.63\% $\pm$ 3.22\% & 52.00\% $\pm$ 3.46\% & 50.70\% $\pm$ 3.10\% & 45.90\% $\pm$ 3.09\%\\
             Swin-T    & 64.80\% $\pm$ 2.96\% & 52.00\% $\pm$ 3.10\% & 48.50\% $\pm$ 3.10\% & 42.70\% $\pm$ 3.07\%\\
             ViT-B-16  & 62.50\% $\pm$ 6.14\% & 51.67\% $\pm$ 6.34\% & 44.27\% $\pm$ 3.52\% & 42.00\% $\pm$ 3.06\%\\
             \midrule
             \citet{liu2023comprehensive} & \textbf{78.00\%} $\pm$ 2.57\% & 59.20\% $\pm$ 3.05\% & 32.50\% $\pm$ 2.90\% & 9.70\% $\pm$ 1.84\%\\
             \citet{debenedetti2023light} & 75.00\% $\pm$ 2.69\% & \textbf{62.30\%} $\pm$ 3.01\% & 17.70\% $\pm$ 2.37\% & 0.70\% $\pm$ 0.52\%\\
             \citet{peng2023robust}       & 71.10\% $\pm$ 2.81\% & 58.20\% $\pm$ 3.06\% & 18.30\% $\pm$ 2.40\% & 1.00\% $\pm$ 0.62\%\\
             \citet{wong2020fast}         & 52.20\% $\pm$ 3.10\% & 33.40\% $\pm$ 2.92\% & 5.10\% $\pm$ 1.36\%  & 0.20\% $\pm$ 0.28\%\\
             \bottomrule
        \end{tabular}
    \end{center}
    \caption{ImageNet top-1 accuracy scores in the black-box, no surrogate setting. The first four rows are standard models to which we added our defense, tuned using APGD at $\varepsilon = 4/255$. The other rows are existing models from RobustBench.}
    \label{tab:square}
\end{table}

\paragraph{Black-box, surrogate.} The results in the black-box, surrogate setting are given in \cref{tab:apgd}. As in the no surrogate setting, we observe a decline in standard accuracy but a significantly higher robust accuracy, especially at large budgets. The best performing model appears to be the WideResNet again. We note that the robust accuracy scores in this setting are lower than the no surrogate setting. This could indicate obfuscated gradients, since transfer attacks should in principle not outperform direct attacks~\citep{carlini2019evaluating,croce2022evaluating,tramer2020adaptive}. We investigate the potential for gradient masking further in \cref{sec:rb_eval}.

\begin{table}[h]
    \begin{center}\footnotesize
        \begin{tabular}{lrrrr}
            \toprule
                    & & \multicolumn{3}{c}{Robust}\\
                    \cmidrule(lr){3-5}
             Model & Standard & $\varepsilon = 8/255$ & $\varepsilon = 16/255$ & $\varepsilon = 32/255$\\
             \midrule
             WRN-101-2 & 70.50\% $\pm$ 2.83\% &  \textbf{37.00\%} $\pm$ 2.99\% & \textbf{37.20\%} $\pm$ 3.00\% & \textbf{37.30\%} $\pm$ 3.00\%\\
             ResNet50  & 70.60\% $\pm$ 2.83\% &  35.60\% $\pm$ 2.97\% & 37.00\% $\pm$ 2.99\% & 34.00\% $\pm$ 2.94\%\\
             Swin-T    & 66.10\% $\pm$ 2.94\% &  29.10\% $\pm$ 2.82\% & 29.80\% $\pm$ 2.84\% & 27.10\% $\pm$ 2.76\%\\
             ViT-B-16  & 56.80\% $\pm$ 3.07\% &  16.10\% $\pm$ 2.28\% & 15.20\% $\pm$ 2.23\% & 14.50\% $\pm$ 2.18\%\\
             \midrule
             \citet{liu2023comprehensive} & \textbf{79.50\%} $\pm$ 2.50\% & 33.20\% $\pm$ 2.92\% &  3.90\% $\pm$ 1.20\% &  0.00\% $\pm$ 0.00\%\\
             \citet{debenedetti2023light} & 73.80\% $\pm$ 2.73\% & 19.30\% $\pm$ 2.45\% &  0.80\% $\pm$ 0.55\% &  0.00\% $\pm$ 0.00\%\\
             \citet{peng2023robust}       & 72.90\% $\pm$ 2.76\% & 27.20\% $\pm$ 2.76\% &  2.60\% $\pm$ 0.99\% &  0.00\% $\pm$ 0.00\%\\
             \citet{wong2020fast}         & 52.70\% $\pm$ 3.10\% & 9.50\% $\pm$ 1.82\%  &  0.40\% $\pm$ 0.39\% &  0.00\% $\pm$ 0.00\%\\
             \bottomrule
        \end{tabular}
    \end{center}
    \caption{ImageNet top-1 accuracy scores in the black-box, surrogate setting. The first four rows are standard models to which we added our defense, tuned using APGD at $\varepsilon = 4/255$. The other rows are existing models from RobustBench.}
    \label{tab:apgd}
\end{table}

We stress that our hyperparameter tuning was conducted at a budget of 4/255, so the defense was not modified in any way to accommodate larger perturbations in the black-box setting. The robustness improvements we obtained therefore seem to generalize across perturbation budgets. By contrast, existing robust models break down very quickly once the budget of 4/255 is exceeded. This is to be expected, since the models we compare against were adversarially trained specifically to be robust at 4/255, so they may outperform our defense if they are re-trained at higher budgets. However, our method is significantly easier to adapt to different budgets, since no training is necessary.

\paragraph{White-box.} The results in the white-box setting are given in \cref{tab:adapt}. Again, the first four rows report the accuracy scores of our defense using different pre-trained base classifiers; the other rows report results of existing work. As in the black-box setting, our best results are obtained using the WideResNet architecture. With high probability, its robust accuracy lies between 47.70\% and 59.24\%, which is very close to the methods of \citet{liu2023comprehensive} and \citet{debenedetti2023light} without the need for adversarial training or data augmentation.

\begin{table}[h]
    \begin{center}
        \begin{tabular}{lrr}
            \toprule
             Model & Standard & $\varepsilon = 4/255$\\
             \midrule
             ResNet50  & 70.60\% $\pm$ 2.83\% & 30.73\% $\pm$ 4.62\%\\
             WRN-101-2 & 70.50\% $\pm$ 2.83\% & 53.47\% $\pm$ 5.77\%\\
             Swin-T    & 66.10\% $\pm$ 2.94\% & 43.38\% $\pm$ 5.90\%\\
             ViT-B-16  & 56.80\% $\pm$ 3.07\% & 27.40\% $\pm$ 6.08\%\\
             \midrule
             \citet{liu2023comprehensive} & \textbf{78.80\%} $\pm$ 2.53\% & \textbf{59.56\%}\\
             \citet{debenedetti2023light} & 75.20\% $\pm$ 2.68\% & 47.60\%\\
             \citet{peng2023robust}       & 71.20\% $\pm$ 2.81\% & 48.94\%\\
             \citet{nie2022diffusion}     & 71.16\% $\pm$ 0.75\% & 44.39\%\\
             \citet{wong2020fast}         & 51.10\% $\pm$ 3.10\% & 26.24\%\\
             \bottomrule
        \end{tabular}
    \end{center}
    \caption{ImageNet top-1 accuracy scores in the white-box setting. The first four rows are standard models to which we added our defense, tuned using APGD at $\varepsilon = 4/255$. The other rows are existing models from RobustBench.}
    \label{tab:adapt}
\end{table}

\subsection{RobustBench evaluation}\label{sec:rb_eval}
Our robustness evaluation follows the guidelines recommended by \citet{croce2022evaluating}. Specifically, our defense can be classified as a randomized input purification method, which is a type of adaptive test-time defense. Notable pitfalls in the robustness evaluation of such methods include:
\begin{itemize}
    \item \textbf{Obfuscated gradients.} Even if the defense is fully differentiable, the iterative nature of the algorithm can cause vanishing or exploding gradients during the backward pass when the number of iterations is high.

    \item \textbf{Randomness.} Like many purification methods, our defense is randomized, which makes the evaluation more difficult and requires us to use EoT~\citep{athalye2018obfuscated} in the white-box setting.

    \item \textbf{Runtime.} In general, iterative purification defenses which require a high number of iterations will naturally reduce inference times, potentially by several orders of magnitude. We do not believe this particular issue pertains to our defense, since the number of iterations we need to perform is very limited compared to most methods: we never exceed 100 iterations, whereas many methods use over 1,000.
\end{itemize}
To address these issues, \citet{croce2022evaluating} recommend evaluating against APGD and Square attacks with a large number of iterations as well as performing transfer attacks. We follow these recommendations here and perform an additional sanity check by verifying that our defense does not degrade robustness of adversarially trained models, which addresses another common pitfall noted by \citet{croce2022evaluating}. The hyperparameters obtained by Optuna for the RobustBench models are given in \cref{tab:rb_parameters}. Notably, these models seem to perform best when using the Fourier transform as opposed to shearlets which are more common for the non-robust models.

\begin{table}
    \begin{center}
        \begin{tabular}{llrrrr}
            \toprule
             Model & Basis & Levels & Threshold & Iterations & Subsampling\\
             \midrule
             \citet{liu2023comprehensive} & Fourier & N/A & 0.11 & 49 & 74.94\%\\ 
             \citet{debenedetti2023light} & DTCWT & 1 & 0.23 & 69 & 74.56\%\\ 
             \citet{peng2023robust} & Fourier & N/A & 0.17 & 59 & 81.31\%\\ 
             \citet{wong2020fast} & Fourier & N/A & 0.16 & 84 & 75.10\%\\ 
             \bottomrule
        \end{tabular}
    \end{center}
    \caption{Hyperparameter settings of the defense for each RobustBench model.}
    \label{tab:rb_parameters}
\end{table}

\Cref{tab:rb_xfer} shows the results of an APGD transfer attack against the RobustBench models with and without our defense. In the Original setting, we directly attack the RobustBench model using APGD without any defense; in the Transfer setting, we evaluate the RobustBench model augmented with our defense on APGD adversarial examples generated in the Original setting. As before, standard accuracy is slightly degraded. However, the robust accuracy is consistently improved by the addition of our defense, although the improvement is not always statistically significant. We conclude that our method does not degrade the robustness of adversarially trained models. The discrepancy in robust accuracy scores between \cref{tab:square,tab:apgd} may therefore be explained by the fact that APGD is a much stronger attack than Square.

\begin{table}[h]
    \begin{center}
        \begin{tabular}{llrr}
            \toprule
            Model & Setting & Standard & $\varepsilon = 4/255$\\
            \midrule
            \citet{liu2023comprehensive}
                & Original & \textbf{79.50\%} $\pm$ 2.50\% & 61.50\% $\pm$ 3.02\%\\
                & Transfer & 75.80\% $\pm$ 2.66\% & \textbf{62.30\%} $\pm$ 3.01\%\\
            \midrule
            \citet{debenedetti2023light}
                & Original & \textbf{73.80\%} $\pm$ 2.73\% & 46.00\% $\pm$ 3.09\%\\
                & Transfer & 62.80\% $\pm$ 3.00\% & \textbf{53.20\%} $\pm$ 3.09\%\\
            \midrule
            \citet{peng2023robust}
                & Original & \textbf{72.90\%} $\pm$ 2.76\% & 50.50\% $\pm$ 3.10\%\\
                & Transfer & 69.50\% $\pm$ 2.86\% & \textbf{53.60\%} $\pm$ 3.09\%\\
            \midrule
            \citet{wong2020fast}
                & Original & \textbf{52.70\%} $\pm$ 3.10\% & 26.30\% $\pm$ 2.73\%\\
                & Transfer & 47.40\% $\pm$ 3.10\% & \textbf{31.40\%} $\pm$ 2.88\%\\
            \bottomrule
        \end{tabular}
    \end{center}
    \caption{ImageNet top-1 accuracy scores for the APGD transfer attack against the RobustBench models.}
    \label{tab:rb_xfer}
\end{table}

\section{Conclusions}
\label{sec:conclusion}
We have shown that the robust width property proposed by \citet{cahill2021robust} can be utilized to derive novel robustness guarantees for any classifier. The resulting defense is lightweight, in the sense that it does not require adversarial retraining or additional data; it requires only the addition of an efficient input purification scheme based on compressed sensing. Our experimental results show that we can significantly outperform state-of-the-art robust models in the black-box setting, especially at large perturbation budgets. In the white-box setting, a careful choice of base classifier allows us to closely match the performance of existing defenses but without the need for expensive adversarial training and data augmentation. We make our code publicly available at \codeurl.

Intriguingly, we find that using convolutional neural networks in conjunction with our defense leads to significantly better results compared to Vision Transformers. Finding an explanation for this disparity may be an interesting avenue for future research, as Vision Transformers are generally deemed to be more robust to adversarial perturbations than convolutional networks~\citep{naseer2021intriguing}. We believe this problem may be attributed to our choice of sensing matrix: we opted for random partial Fourier matrices in this work since they satisfy the theoretical requirements and are efficient to compute. However, the Fourier transform may be unsuitable for certain model architectures such as the Transformer.

The bounds we obtained here are directly based on the results by \citet{cahill2021robust}. However, should better bounds become available, our framework is sufficiently general to readily incorporate them. In particular, the current state of the art in compressed sensing based on robust width seems to require an inherent robustness of at least $2\varepsilon/\alpha$, which means we need RWP operators with $\alpha > 2$ in order to have non-trivial robustness gain. An interesting open question arising from this work is whether this bound is tight. While it is easy to show that non-zero robustness of the base classifier is necessary in general, the question remains how much robustness is truly required. We conjecture that any lower bound based on robust width will satisfy $\Omega(\varepsilon/\alpha)$, but the constants involved may be significantly improved.

This work focuses on improving adversarial robustness in resource-limited environments. As such, our defense does not make use of additional models and instead relies only on ``classical'' algorithms. However, in recent years connections between compressed sensing and robustness in deep neural networks have been explored~\citep{bora2017compressed,mukherjee2023learned}. In particular, the denoising diffusion restoration models proposed by \citet{kawar2022denoising} may be used to further improve robustness in settings where resources are not limited and the use of large diffusion models is viable. Within the setting of ADMM plug-and-play denoising methods, such models may be incorporated and/or combined with our defense method. Applying our theoretical results in this setting and empirically evaluating the robustness of the resulting models is another interesting line of research.

The use of the RWP yields an interesting theoretical robustness bound (\cref{thm:adv}) which is in some sense more interpretable than the bound \eqref{eq:dsrob} of the DS algorithm. Indeed, our result depends on the assumption that the data manifold is characterized by vectors which are sparse in some appropriate basis. Under such a sparsity assumption, we can give meaningful robustness bounds for any classifier that depend linearly on the distance of the given samples to this manifold. This is a powerful result, since it is believed that empirical data tend to obey the so-called \textit{manifold hypothesis}, \ie the idea that the data are concentrated on a low-dimensional smooth manifold embedded in $\mathbb R^n$~\citep{donoho2000high,fefferman2016testing}. This hypothesis is a necessary condition for most machine learning algorithms to work. Our robustness bound can be interpreted as a special case of a more general result which holds for any smooth manifold. Generalizing our results to arbitrary smooth manifolds would be a very interesting avenue for future work. 
 
\section*{Acknowledgements}
This research received funding from the Flemish Government under the ``Onderzoeksprogramma Artifici\"ele Intelligentie (AI) Vlaanderen'' programme.

\bibliography{main}

\end{document}